\documentclass[runningheads]{llncs}
\usepackage{graphicx} \graphicspath{ {figure/} }
\usepackage{comment}
\usepackage{amsmath,amssymb,mathabx} 
\usepackage{algorithmic}
\usepackage[ruled]{algorithm2e}
\usepackage{acronym}
\usepackage{enumitem}
\usepackage{hyperref}
\usepackage{balance}
\usepackage{xspace}
\usepackage{setspace}
\usepackage{subcaption}
\usepackage{caption}
\usepackage[misc]{ifsym}
\usepackage[dvipsnames]{xcolor}
\usepackage[capitalise]{cleveref}
\usepackage{tabularx}
\usepackage{multirow,multirow,array,makecell}
\usepackage{array}
\usepackage{makecell}
\usepackage{overpic}
\usepackage{wrapfig}
\usepackage{pifont}
\usepackage{booktabs}

\makeatletter
\DeclareRobustCommand\onedot{\futurelet\@let@token\@onedot}
\def\@onedot{\ifx\@let@token.\else.\null\fi\xspace}
\def\eg{\emph{e.g}\onedot} 

\def\ie{\emph{i.e}\onedot}

\def\etal{\emph{et al}\onedot}
\makeatother

\newcommand{\cmark}{\ding{51}}
\newcommand{\xmark}{\ding{55}}

\makeatletter
\def\BState{\State\hskip-\ALG@thistlm}
\makeatother

\begin{document}
\pagestyle{headings}
\mainmatter
\def\ECCVSubNumber{5581}  
\titlerunning{A Multi-view Dataset for \underline{L}\underline{E}arning \underline{M}ulti-agent \underline{M}ulti-task \underline{A}ctivities}

\title{LEMMA: A Multi-view Dataset for \underline{L}\underline{E}arning \underline{M}ulti-agent \underline{M}ulti-task \underline{A}ctivities}

\author{Baoxiong Jia\and
Yixin Chen \and
Siyuan Huang \and
Yixin Zhu \and 
Song-Chun Zhu
}
\authorrunning{Baoxiong Jia et al.}

\institute{UCLA Center for Vision, Cognition, Learning, and Autonomy (VCLA)\\
\email{\{baoxiongjia, ethanchen, huangsiyuan, yixin.zhu\}@ucla.edu} \email{sczhu@stat.ucla.edu}}

\maketitle

\begin{abstract}
Understanding and interpreting human actions is a long-standing challenge and a critical indicator of perception in artificial intelligence. However, a few imperative components of daily human activities are largely missed in prior literature, including the goal-directed actions, concurrent multi-tasks, and collaborations among multi-agents. We introduce the LEMMA dataset to provide a single home to address these missing dimensions with meticulously designed settings, wherein the number of tasks and agents varies to highlight different learning objectives. We densely annotate the atomic-actions with human-object interactions to provide ground-truths of the compositionality, scheduling, and assignment of daily activities. We further devise challenging compositional action recognition and action/task anticipation benchmarks with baseline models to measure the capability of compositional action understanding and temporal reasoning. We hope this effort would drive the machine vision community to examine goal-directed human activities and further study the task scheduling and assignment in the real world.
\keywords{Dataset, Multi-agent Multi-task Activities, Compositional Action Recognition, Action and Task Anticipations, Multiview}
\end{abstract}

\section{Introduction}

Activity understanding is one of the most fundamental problems in artificial intelligence and computer vision. As the most readily available learning source, videos of daily human activities could be used to train intelligent agents and, in turn, to assist humans. However, compared to recent progress in learning from static images~\cite{antol2015vqa,he2016deep,he2017mask,ren2015faster}, current machine vision's ability to understand activities from videos still falls short. Admittedly, activity understanding is inherently more challenging, which requires reason about the complex structures in activities along the additional temporal dimension; but we argue there are more profound reasons that we must look back to the origin of activity understanding.

\begin{figure}[t!]
    \centering
    \includegraphics[width=\linewidth]{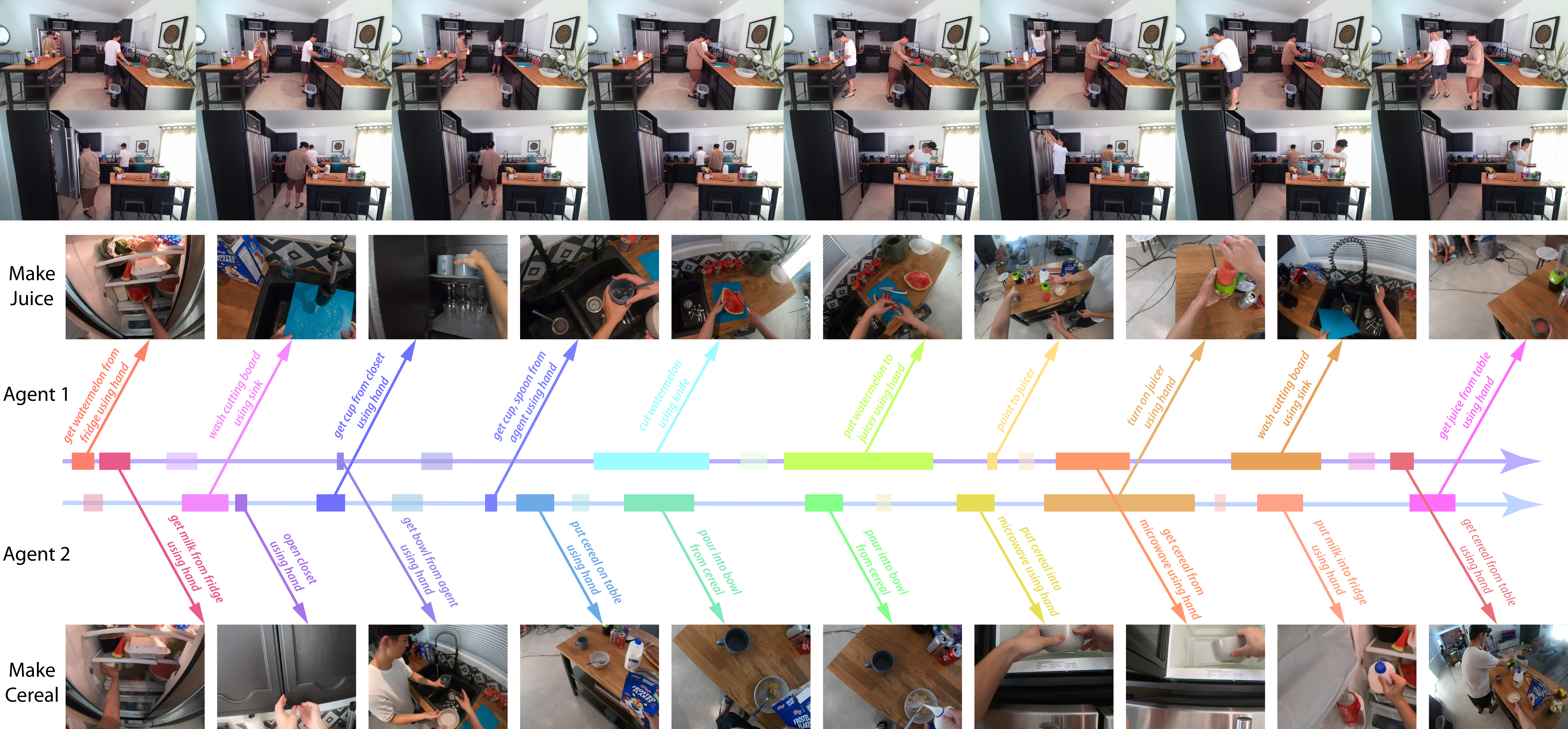}
    \caption{Illustrations of the proposed multi-view dataset with annotations. From top to bottom: frames captured from the third-person primary view, frames captured from the third-person side view, annotated segments of each agent executing tasks, and corresponding frames captured from the first-person view.}
    \label{fig:overview}
\end{figure}

The study and analysis of human motion perception are rooted in the field of neuroscience~\cite{turaga2008machine}. Using a dot-representation of human motions, Johansson~\cite{johansson1973visual} adopted a method to produce proximal patterns (\ie, the moving light display experiment), which demonstrated that human perception of activities does not tightly couple with \emph{pixel-based features}; human subjects can still perceive the semantics of activities from \emph{sparse} representations of motions. Evidence from developmental psychology, the classic Heider-Simmel experiment, further suggests that we perceive human activities from as \emph{goal-directed} behaviors~\cite{woodward1998infants,baldwin2001infants,gergely2002rational,csibra2007obsessed}; it is the underlying intent, rather than the surface pixels or behavior, that matters when we observe motions~\cite{baldwin2001discerning}. Such a \textbf{goal-directed~\cite{land1999roles} perspective} of activity understanding has been largely left untouched in computer vision.

Daily human activities are intrinsically multi-tasked~\cite{monsell2003task,rubinstein2001executive}; understanding activity naturally demands a learning system to interpret concurrent interactions. As agents' decision-making processes are deeply affected by their unique social values, task scheduling is significantly affected by interactions (\eg, cooperation, competition, subordination) among multi-agents~\cite{kleiman2016coordinate}. These observations implicate that the machine vision system must objectively understand how a given task should be decomposed into atomic-actions, how multi-tasks should be executed and coordinated in parallel among multi-agents, and take the perspective from human agents to understand why the observed human activities are optimal solutions. Such a \textbf{decompositional, multi-task, multi-agent, diagnostic-driven, social perspective} of activity understanding is critical for an intelligent agent to understand human behavior and team with humans collaboratively; yet it is broadly missing in activity understanding literature.

The semantics of human actions are intrinsically ambiguous when described in natural language. For instance, although both ``opening the fridge'' and ``opening a book'' use the action verb ``open,''  their semantics of the actions are utterly different. In this paper, we take the stance of Grice's influential work on language act~\cite{grice1975logic}---technical tools for reasoning about rational action should elucidate linguistic phenomena~\cite{goodman2016pragmatic}. Specifically, the compositional relations between the verbs and nouns could reveal the functionality of the object and the patterns of human-object interactions, which subsequently facilitate the understanding of the observed human activities and the language that describes them. Though the previous work~\cite{goyal2017something} attempted to address this issue, more general and flexible \textbf{compositional relations for describing human actions interacting with objects} are requisite for a goal-directed activity understanding.

Motivated by these deficiencies in prior work, we introduce the LEMMA dataset to explore the essence of complex human activities in a goal-directed, multi-agent, multi-task setting with ground-truth labels of compositional atomic-actions and their associated tasks. By quantifying the scenarios to up to two multi-step tasks with two agents, we strive to address human multi-task and multi-agent interactions in four scenarios: single-agent single-task ($1 \times 1$), single-agent multi-task ($1 \times 2$), multi-agent single-task ($2 \times 1$), and multi-agent multi-task ($2 \times 2$). Task instructions are only given to one agent in the $2 \times 1$ setting to resemble the robot-helping scenario, hoping that the learned perception models could be applied in robotic tasks (especially in HRI) in the near future.

Both the third-person views (TPVs) and the first-person views (FPVs) were recorded to account for different perspectives of the same activities; see \cref{fig:overview}. Such a setting potentially benefits future study on 3D holistic scene understanding~\cite{chen2019holistic++,huang2018holistic}, as well as action understanding and prediction~\cite{ryoo2011human,qi2020generalized}. We densely annotate atomic-actions (in the form of compositional verb-noun pairs) and tasks of each atomic-action, to facilitate the learning of multi-agent multi-task task scheduling and assignment; see more details in \cref{sec:dataset}.

\subsection{Related Work}

In this section, we review and compare prior indoor activity datasets on the basis of tasks and captured video contents; see a detailed summary in \cref{tab:dataset_comparison}.

\begin{table}[b!]
    \centering
    \caption{Comparisons between LEMMA and relevant indoor activity datasets.}
    \label{tab:dataset_comparison}
    \resizebox{\hsize}{!}{%
    \begin{tabular}{cccccccccccc}
        \toprule
        \multirow{2}{*}{\shortstack[c]{Dataset}} & \multirow{2}{*}{\shortstack[c]{Task\\ Annotation}} & 
        \multirow{2}{*}{\shortstack[c]{Multi- \\ agent}} & \multirow{2}{*}{\shortstack[c]{Multi-\\task}} & \multirow{2}{*}{\shortstack[c]{Multi-\\view}} &
        \multirow{2}{*}{\shortstack[c]{Samples}}
        & \multirow{2}{*}{\shortstack[c]{Frames}} & \multirow{2}{*}{\shortstack[c]{Action\\Classes}} & \multirow{2}{*}{\shortstack[c]{Action\\Segments}} & \multirow{2}{*}{\shortstack[c]{Actions per\\Video}} & \multirow{2}{*}{\shortstack[c]{Modality}} & \multirow{2}{*}{\shortstack[c]{Year}}\\
        & & & & & & & \\
        \hline
        MPII Cooking~\cite{rohrbach2012database} & \cmark & \xmark & \xmark & \xmark & 273 & 2.9M  & 88 & 14,105& 51.7 & RGB & 2012\\
        ADL~\cite{pirsiavash2012detecting} & \xmark & \xmark & \cmark & \xmark & 20 & 1.0M & 32 & 436 & 13.6 & RGB & 2012\\
        50Salads~\cite{stein2013combining} & \cmark & \xmark & \xmark & \xmark & 50 & 0.5M  & 17 & 966 & 19.3 & RGB-D & 2013\\
        CAD-120~\cite{koppula2013learning} & \xmark & \xmark & \xmark & \xmark & 120 & 0.1M & 10 & 1,175 & 9.8 & RGB-D & 2013\\
        Breakfast~\cite{kuehne2014language} & \cmark &  \xmark & \xmark & \cmark & 433 & 3.0M & 50 & 3,078  & 7.1 & RGB & 2014\\
        Watch-n-Patch~\cite{wu2015watch} & \cmark & \xmark & \xmark & \xmark & 458 & 0.1M & 21 &  2978 & 6.5 & RGB-D & 2015\\
        Charades~\cite{sigurdsson2016hollywood} & \xmark & \xmark & \cmark & \xmark & 9,848 & 7.4M & 157 & 67,000  & 6.8 & RGB & 2016\\
        Something-Something~\cite{goyal2017something} & \xmark & \xmark & \xmark & \xmark & 108,499 & - & 174 & 108,499 & 1.0 & RGB & 2017\\
        EGTEA GAZE+~\cite{li2018eye} & \cmark & \xmark & \xmark & \xmark & 86 & 2.4M & 106 & 10,325 & 120.1 & RGB & 2018\\
        EPIC-KITCHENS~\cite{damen2018scaling} & \xmark & \xmark & \cmark & \xmark & 432 & 11.5M & 149  & 39,596 & 91.7 & RGB & 2018\\
        \hline
        LEMMA (proposed) & \cmark & \cmark & \cmark & \cmark & 324 & 4.6M & 641  & 11,781 & 36.4 & RGB-D & 2020\\
        \bottomrule
        \end{tabular}
    }
\end{table}

Crowd-sourced from online videos and movie sharing platforms, typical large-scale video datasets~\cite{soomro2012ucf101,karpathy2014large,caba2015activitynet,carreira2017quo,fouhey2018lifestyle} focus on \textbf{video-level summarization and classification}. Although activity classes exhibit a large inter-class variability, spanning from outdoor sports activities to indoor household activities, they generally lack sequential, goal-directed activities. Notably, they suffer from a major drawback~\cite{girdhar2019cater}; activities are highly correlated to the general scene and object context, possessing a strong dataset bias for activity understanding.

Some datasets tackle the \textbf{human atomic-actions} using short clips or limited tasks, with a focus on the semantics of action verbs and objects~\cite{goyal2017something}, 3D action analysis~\cite{li2010action,ionescu2013human3,savva2016pigraphs}, and action grounding with multi-modality inputs~\cite{monfort2019moments}. Although such datasets are suitable for atomic-actions, they are intrinsically impaired at studying the long-term reasoning of goal-directed human activities.

Recently, \textbf{concurrent actions} have been taken into consideration. For instance, Charades~\cite{sigurdsson2016hollywood} is a large-scale benchmark for household activities, and Charades-Ego~\cite{sigurdsson2018charades} steps further with both FPVs and TPVs. However, the activities involved are mostly unrelated to specific goals due to the crowdsourced script generation process. Similarly, although Multi-THUMOS~\cite{yeung2018every} and AVA~\cite{gu2018ava} focus on highly paralleled activities, and some datasets look at the temporal order of activities~\cite{bojanowski2014weakly,tapaswi2016movieqa}, the unnaturally scripted activities result in the lack of meaningful goal-directed tasks exhibited in our daily life.

Conversely, \textbf{instructional video} datasets~\cite{alayrac2016unsupervised,stein2013combining,kuehne2014language,koppula2013learning,rohrbach2016recognizing} tackle goal-directed multi-step tasks, mostly in cooking, repairing, and assembling activities. In spite of their relevance, they fail to account for multi-agent or multi-task problems. EPIC-KITCHENS~\cite{damen2018scaling} is perhaps the only exception; it records naturally paralleled task execution of agents in kitchen environments, but with no task specification or multi-agent interactions. Additionally, prior instructional video datasets have either drastic view perspective changes~\cite{zhou2018towards,alayrac2016unsupervised,tang2019coin,toyer2017human} or limited egocentric view with severe occlusions~\cite{pirsiavash2012detecting,li2018eye}, hindering the activity understanding.

Another related stream of work is the learning of group-level activities in a \textbf{multi-agent} setting~\cite{ibrahim2016hierarchical}, such as detecting key actors~\cite{ramanathan2016detecting}, predicting future trajectories~\cite{pellegrini2009you,lerner2007crowds}, and recognizing collective activities~\cite{choi2009they,oh2011large,shu2015joint}. However, such coarse-grained multi-agent interactions leave the latent subtlety of collaboration and task assignment untouched. Although simulation-based multi-agent environments~\cite{baker2019emergent,vinyals2019grandmaster,berner2019dota} can partially address such an issue, learning from noisy and real visual input in physical work is still essential for understanding collaborative planning behaviors of agents in the context of complex daily tasks.

The collected LEMMA dataset strives to address the shortcomings of the aforementioned works, capturing goal-directed, decompositional, multi-task activities with multi-agent collaborations. As shown in \cref{tab:dataset_comparison}, the size, annotation, and actions per video of LEMMA are at a comparable scale to state-of-the-art benchmarks. We hope such a design will boost the study of human activity understanding and potentially motivate new cross-disciplinary research insights.

\subsection{Contributions}

This paper's contribution is three-fold. (i) We design and collect a multi-view video dataset, capturing multi-agent, multi-task activities with goal-directed daily tasks. (ii) We annotate the dataset, focusing on the compositionality of actions and the governing task for each atomic-action. (iii) We provide compositional action recognition and action/task anticipation benchmarks by considering the aforementioned features; we also compare and analyze multiple baseline models to promote future research on human activity understanding.

\section{The LEMMA Dataset}\label{sec:dataset}

This section describes the design, data collection, and data annotation process of the LEMMA dataset. The dataset is profiled by various statistics from diversified perspectives to highlight its potentials in activity understanding.\footnote{The dataset will be made publicly available at the following website with download links and util code: \url{https://sites.google.com/view/lemma-activity}.}

\subsection{Activities and Scenarios}

We first build a task pool of 15 common tasks in the kitchen (\eg, ``make juice,'' ``make cereal'') and living room (\eg ``watch TV,'' ``water plant''). On top of these tasks, we design four types of scenarios (with a different focus) to study goal-directed multi-step multi-task indoor activities in multi-agent settings.
\begin{enumerate}[leftmargin=*,noitemsep,nolistsep,wide=0pt]
    \item\textbf{Single-agent Single-task ($1 \times 1$):}
Each participant was first asked to perform all tasks from the task pool independently; this ensures participants are clear with the goal of each task and could schedule and assign tasks efficiently in later multi-task or multi-agent scenarios. Participants were asked to read the instructions and walk around to get familiarized with the new environments.
    \item\textbf{Single-agent Multi-task ($1 \times 2$):}
Each participant was then asked to simultaneously perform two tasks, randomly sampled from the task pool. The participants determined the order of task executions without any restrictions.
    \item\textbf{Multi-agent Single-task ($2 \times 1$):}
Two participants were asked to perform a single task cooperatively; the task is randomly selected from the task pool. To emulate human-robot teaming accurately, only one participant (leader) was provided with task instructions; the other participant (helper), with no knowledge of the task, was asked to collaborate with the leader agent to finish the task efficiently. Only nonverbal communications (\eg, gestures) were allowed between two participants; this design would open up new venues on nonverbal communications and the emergence of language in real-world environments.
    \item\textbf{Multi-agent Multi-task ($2 \times 2$):}
Both participants were provided with task instructions. Since both participants were asked to accomplish two complex multi-step tasks collaboratively, this scenario has the most natural activity/task patterns and richest mechanisms for learning task scheduling and assignment.
\end{enumerate}

\begin{figure}[t!]
    \centering
    \includegraphics[width=\linewidth]{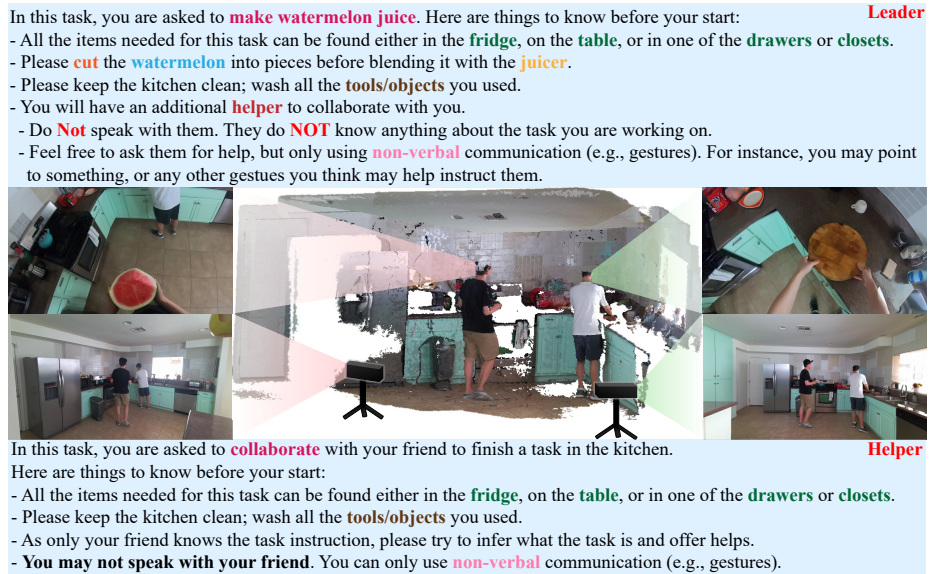}
    \caption{An exemplar task instruction of making juice for two agents in a \textbf{Multi-agent Single-task ($2 \times 1$)} scenario. Middle: Point clouds, TPVs, and FPVs.}
    \label{fig:data_collection}
\end{figure}

In total, the LEMMA dataset includes 37 unique task combinations in the multi-task scenarios. Participants were explicitly instructed to perform tasks efficiently and provided with a brief task instruction with basic environment information. Except for the specification of the goal states for each task, we add no additional constraint to the order of task execution; participants perform tasks naturally and freely. \cref{fig:data_collection} shows a sample instruction for the $2 \times 1$ scenario.

\subsection{Data Collection}

We recorded the data in 7 different Airbnb houses, performed by 8 individuals in 14 unique kitchens/living rooms. To provide different views of performing the daily activities and avoid occlusion in narrow spaces, we set up two Kinect Azure cameras to capture the RGB-D videos of the global scene and human bodies. In addition, each participant was instructed to wear a head-mounted GoPro camera to capture detailed agent-specific actions in an egocentric view. 
In post-processing, we synchronize the camera recordings of all views at a frame rate of 24 FPS. \cref{fig:data_collection} shows an example of a scene with a point cloud merged from two Kinects and four RGB views from both Kinects and GoPros. Combining TPVs and FPVs captures most of the details of performing daily activities, provides sufficient data for understanding human activities, and benefits future research in embodied vision. The additional depth information and 3D human skeletons captured by Kinects can also be adopted for future 3D understanding tasks. 

\subsection{Ground-truth Annotation}

We used the Amazon Mechanical Turk (AMT) to annotate both human bounding boxes and action information in the synchronized recordings. Specifically, action information includes the temporal localization of segments, semantic labels, and the governing task of each atomic-action. The semantic labels of atomic-actions are composed of verbs and nouns, representing flexible compositional relations to describe human actions. Additional details are provided below.

\paragraph{\textbf{Bounding Boxes and Segments:}}
Bounding boxes of humans are annotated on the primary view of TPVs. Skeletons captured by Kinects are used to provide initial estimations of bounding boxes. Next, we use Vatic~\cite{vondrick2013efficiently} to adjust bounding boxes and annotate the segments of atomic-actions. The segments of atomic-actions are defined by verbs without corresponding nouns, for example, ``put \underline{\ \ \ } to \underline{\ \ \ } using \underline{\ \ \ },'' ``pour into \underline{\ \ \ } from \underline{\ \ \ }.'' Each video was first annotated by two AMT workers; task-irrelevant actions (\eg, ``walking,'' ``holding'') are ignored. We then compute the Intersection over Union (IoU) of both bounding boxes and temporal segments. A third AMT worker is asked to fine-tune the annotations if the IoU of bounding boxes or segments annotated is lower than 0.5.

\begin{figure}[t!]
    \centering
	\begin{subfigure}[t]{\linewidth}
		\includegraphics[width=\linewidth,trim={2.8cm 0.8cm 0.3cm 0.3cm},clip]{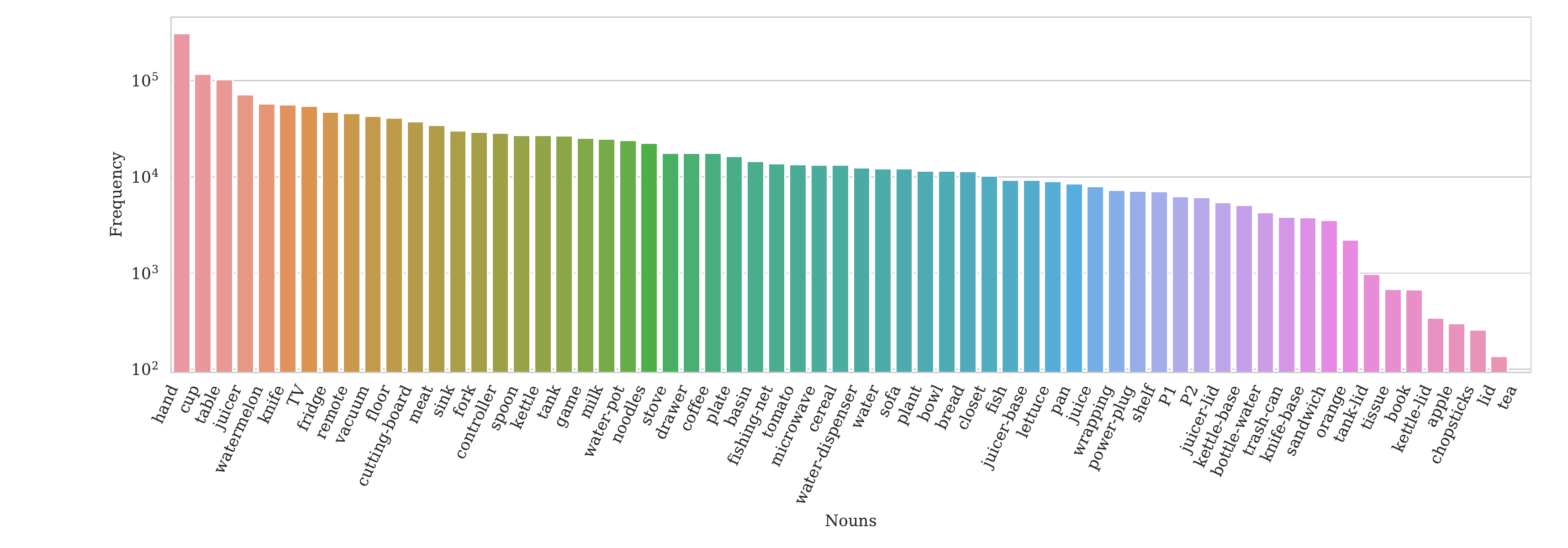}
		\caption{Frequency of annotated noun classes across all frames}
		\label{fig:data_statistics_noun}
	\end{subfigure}%
	\\
	\begin{subfigure}[t]{0.375\linewidth}
		\includegraphics[width=\linewidth,trim={0.8cm 0.8cm 0.3cm 0.3cm},clip]{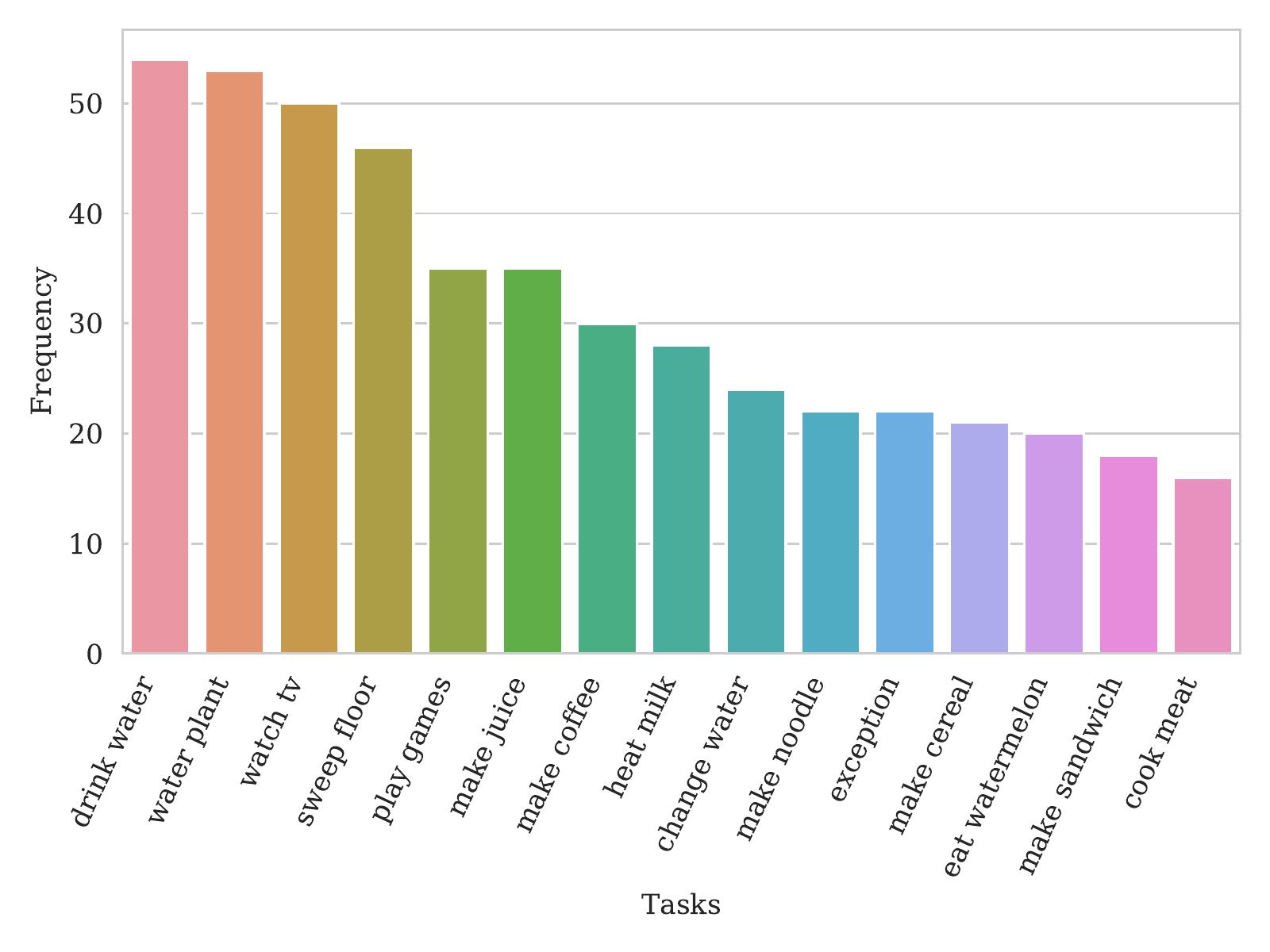}
		\caption{Frequency of recorded tasks}
		\label{fig:data_statistics_task}
	\end{subfigure}%
	\begin{subfigure}[t]{0.375\linewidth}
		\includegraphics[width=\linewidth,trim={0.8cm 0.8cm 0.3cm 0.3cm},clip]{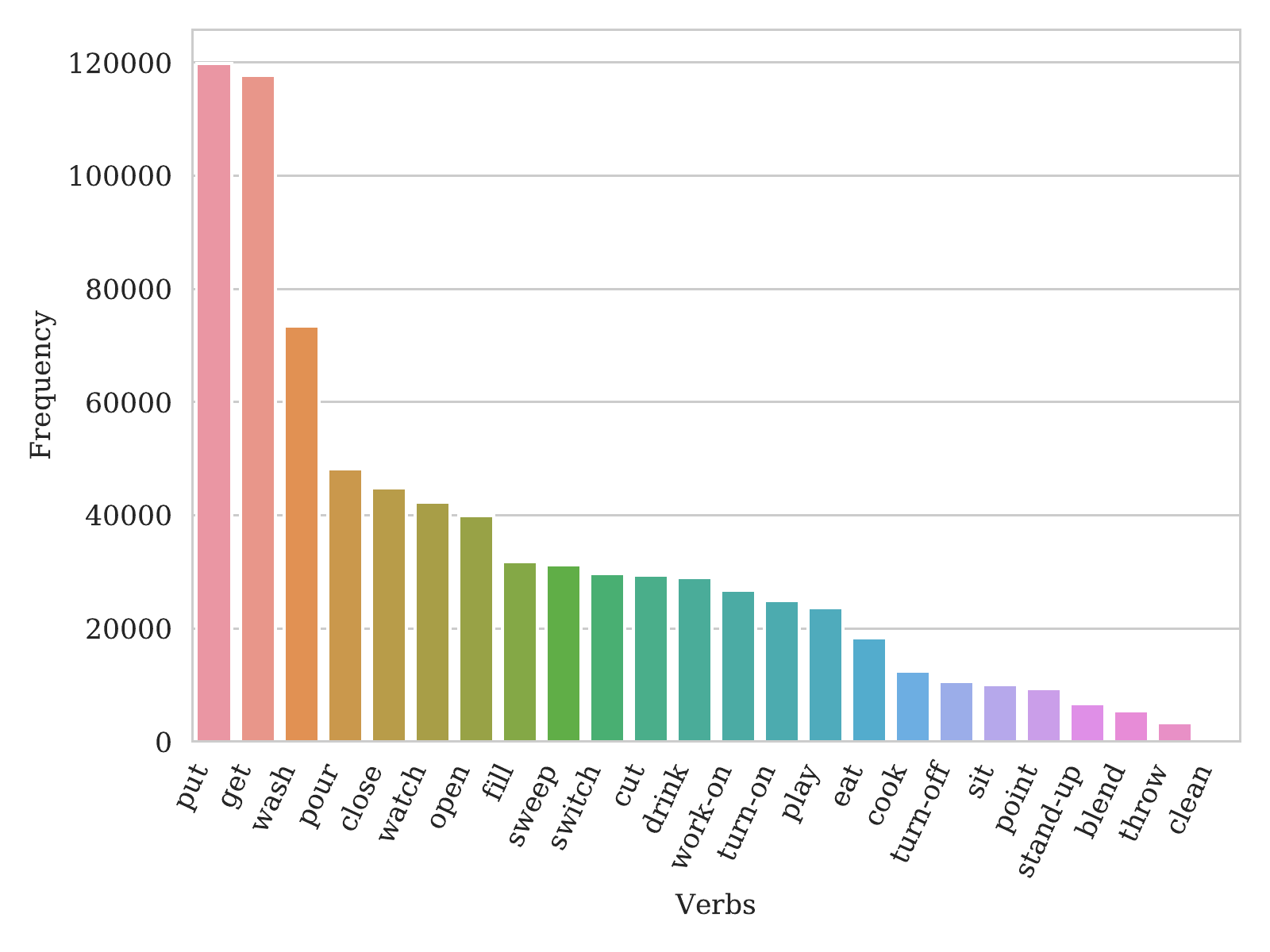}
		\caption{Frequency of annotated verb}
		\label{fig:data_statistics_verb}
	\end{subfigure}%
	\begin{subfigure}[t]{0.25\linewidth}
		\includegraphics[width=\linewidth,trim={0cm -1.6cm 0cm 0cm},clip]{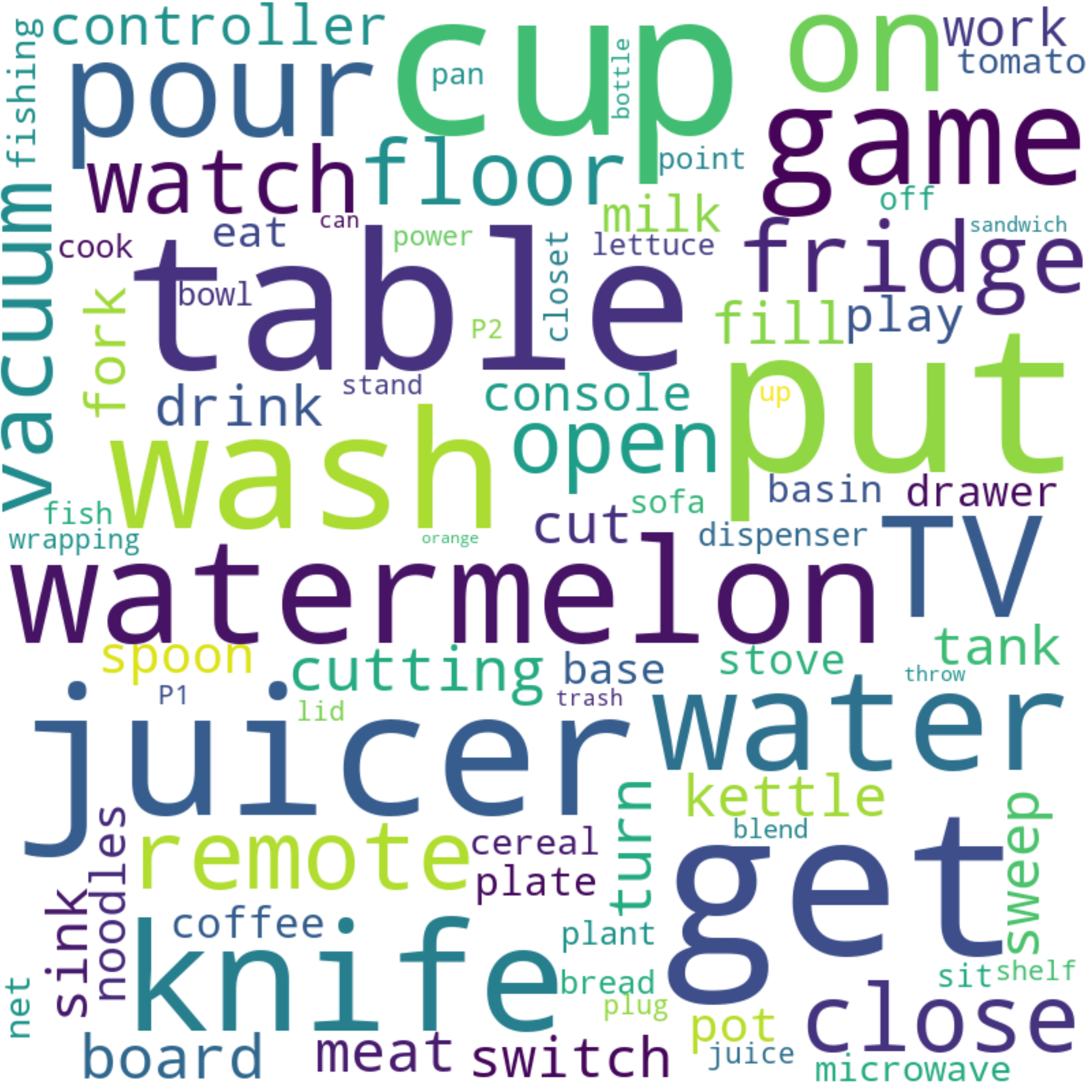}
		\caption{Action wordle}
		\label{fig:data_statistics_wordle}
	\end{subfigure}%
    \caption{Statistics of the LEMMA dataset.}
    \label{fig:data_statistics}
\end{figure}

\paragraph{\textbf{Atomic-actions and Activities:}}
Given the verbs of the atomic-action segments, two AMT workers were asked to fill in the blanks of the verb patterns and annotate the governing tasks in multi-task scenarios with a self-developed interactive annotation tool (see \emph{supplementary material}). We allow concurrent actions for each agent with multiple nouns for the same verb; for example, ``get \underline{spoon, cup} from \underline{table} using \underline{hand}.'' As there might exist ambiguities in describing the atomic-actions with natural languages, such as the possible annotations of ``wash \underline{cup} using \underline{water}'' \emph{vs.} ``wash \underline{cup} using \underline{sink},'' we manually go through all the annotations and resolve the ambiguous action annotations following a uniform criterion. Examples of annotation results are shown in \emph{supplementary}.

\begin{figure}[t!]
    \centering
	\begin{subfigure}[t]{0.445\linewidth}
		\includegraphics[width=\linewidth,trim={0.5cm 2.5cm 0.6cm 0.6cm},clip]{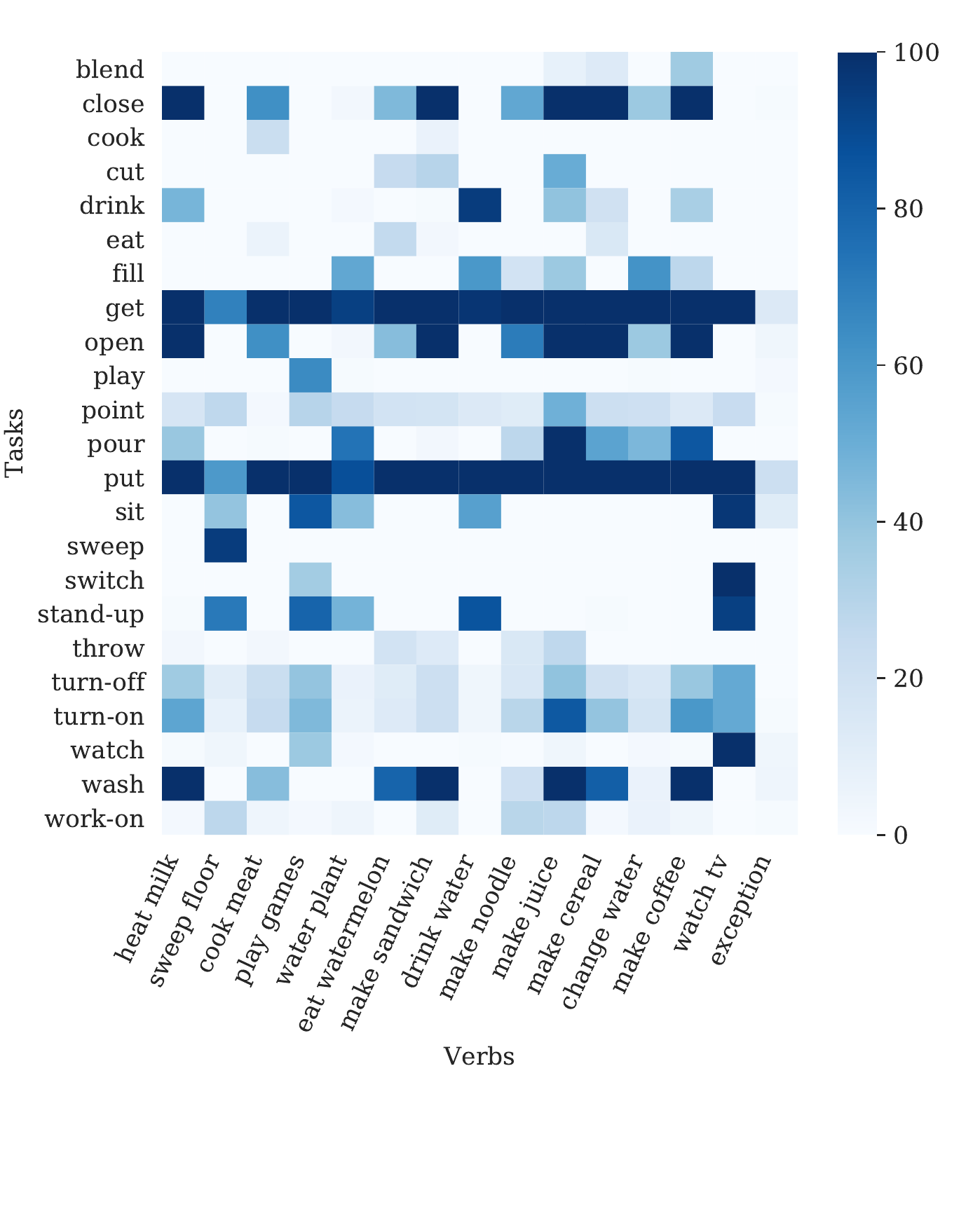}
		\caption{verbs (y-axis) and tasks (x-axis)}
		\label{fig:data_cooccur_verb_task}
	\end{subfigure}%
    \begin{subfigure}[t]{0.555\linewidth}
		\includegraphics[width=\linewidth,trim={0.8cm 0.8cm 3.1cm 0.4cm},clip]{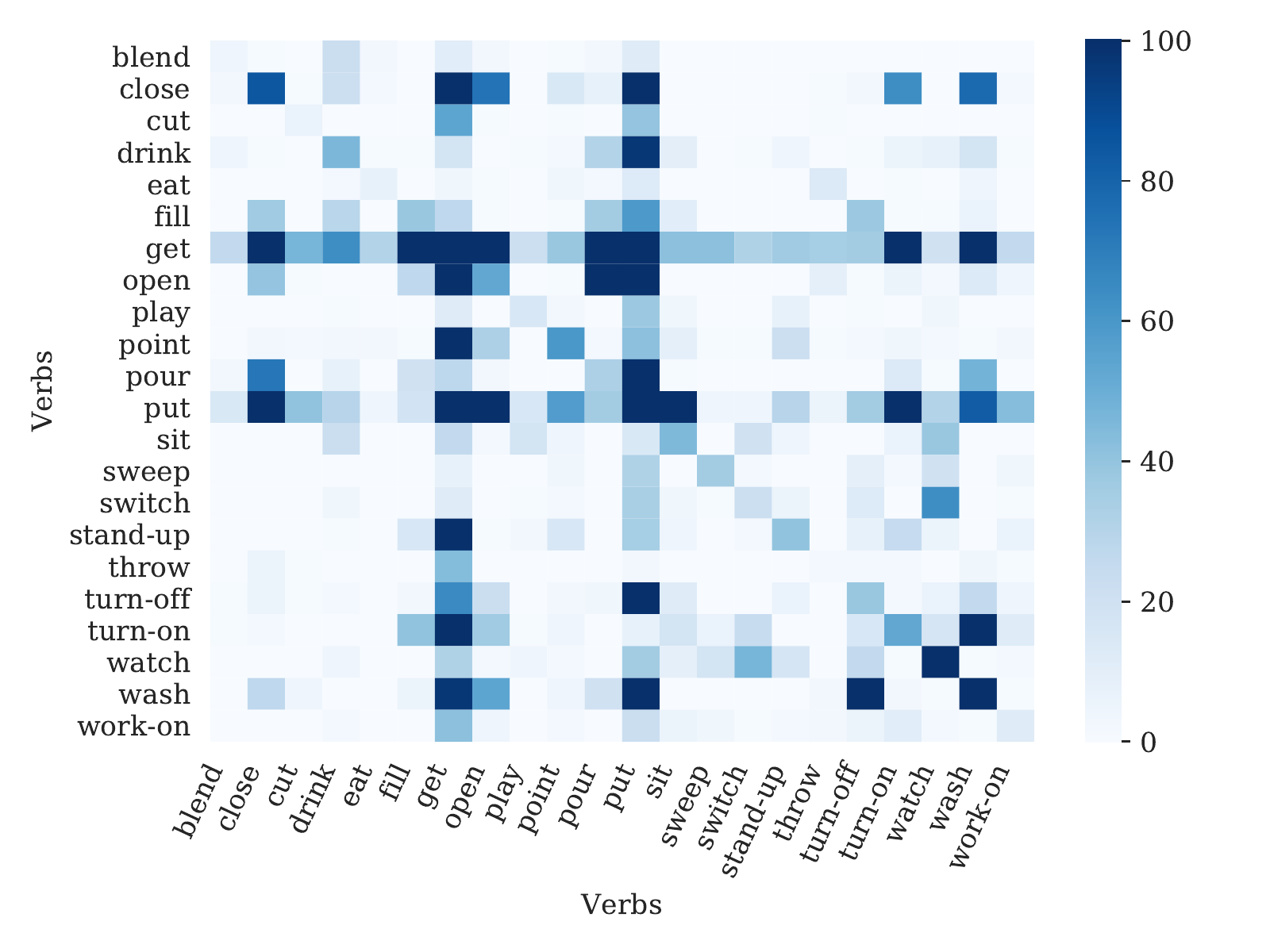}
		\caption{verbs (y-axis) and next verbs (x-axis)}
		\label{fig:data_cooccur_verb_verb}
	\end{subfigure}%
	\\
	\begin{subfigure}[t]{\linewidth}
		\includegraphics[width=\linewidth,trim={4.6cm 0.6cm 7.5cm 0.3cm},clip]{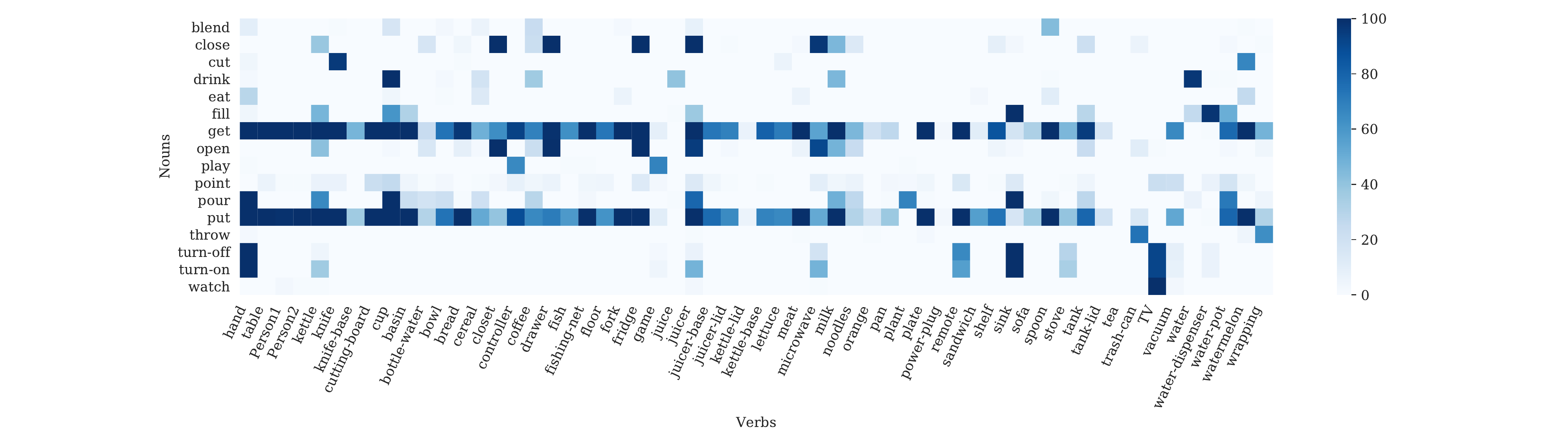}
		\caption{verbs (y-axis) and nouns (x-axis)}
		\label{fig:data_cooccur_verb_noun}
	\end{subfigure}%
    \caption{The co-occurrence statistics for verbs, nouns, and tasks in LEMMA.}
    \label{fig:data_cooccur}
\end{figure}

\subsection{Dataset Statistics}

In total, we recorded 324 activities, generating $324 \times 2$ TPV videos (from both Kinects) and 445 FPV videos. Among them, 136 activities were performed in kitchens and the remaining 188 in the living rooms. The collected LEMMA dataset consists of 127 $1 \times 1$ activities, 76 $1 \times 2$ activities, 66 $2 \times 1$ activities, and 55 $2 \times 2$ activities. The frequency of the recorded tasks is shown in \cref{fig:data_statistics_task}. The total duration of all the activities is 10.1 hours, with an average duration of 2 minutes per video and the longest activity of 7 minutes.

We retrieved a total of 4.6 million images during post-processing, including 2.9 million RGB images captured by both GoPros and Kinects and 1.7 million depth images captured by Kinects. We annotated 0.9 million RGB frames captured by the primary view Kinect and gathered 0.8 million annotated frames with one or more actions performed by each of the agents (if multiple).

After resolving annotation ambiguities, we collected 24 verb classes and 64 noun classes, resulting in 862 compositional atomic-action labels, of which 641 appear more than 50 times. We show the frequencies of annotated verbs and nouns in \cref{fig:data_statistics_noun,fig:data_statistics_verb}; both distributions roughly follow the Zipf's law.

Co-occurrence relations among annotated verbs, nouns, and tasks are shown in \cref{fig:data_cooccur}. As we can see from \cref{fig:data_cooccur_verb_task,fig:data_cooccur_verb_noun}, verbs like ``get'' and ``put'' co-occur with various nouns in almost all of the tasks, which aligns with our intuition that moving objects around consists a large portion of our daily activities. Interactive actions between participants are captured by verbs (\eg, ``point-to'') and nouns (\eg, ``P1,'' short for ``participant 1'') in the form of annotations like ``get \underline{knife} from \underline{P1} using \underline{hand}'' or ``point-to \underline{sink}.''

\section{Benchmarks}

Aligned with our motivations, two general goals are constructed to evaluate indoor human activity understanding on the collected LEMMA dataset: (i) recognize atomic-actions and their semantics; and (ii) understand the goal-directed activities and monitor multiple concurrent tasks, especially in multi-agent scenarios. Specifically, we define two challenging benchmarks to test the capability of understanding complex goal-directed activities for computer vision algorithms.

\subsection{Compositional Action Recognition}

Human indoor activities are composed of fine-grained action segments with rich semantics. As mentioned by Goyal \etal~\cite{goyal2017something}, interactions with objects are highly purposive. From the simplest verb of ``put,'' we can generate a plethora of combinations of objects and target places, such as ``put \underline{cup} onto \underline{table},'' ``put \underline{fork} into \underline{drawer}.'' Situations could become even more challenging when objects were used as tools; for example, ``put \underline{meat} into \underline{pan} using \underline{fork}.''

\begin{figure*}[t!]
    \centering
    \begin{subfigure}[t]{0.46\linewidth}
        \includegraphics[width=\linewidth]{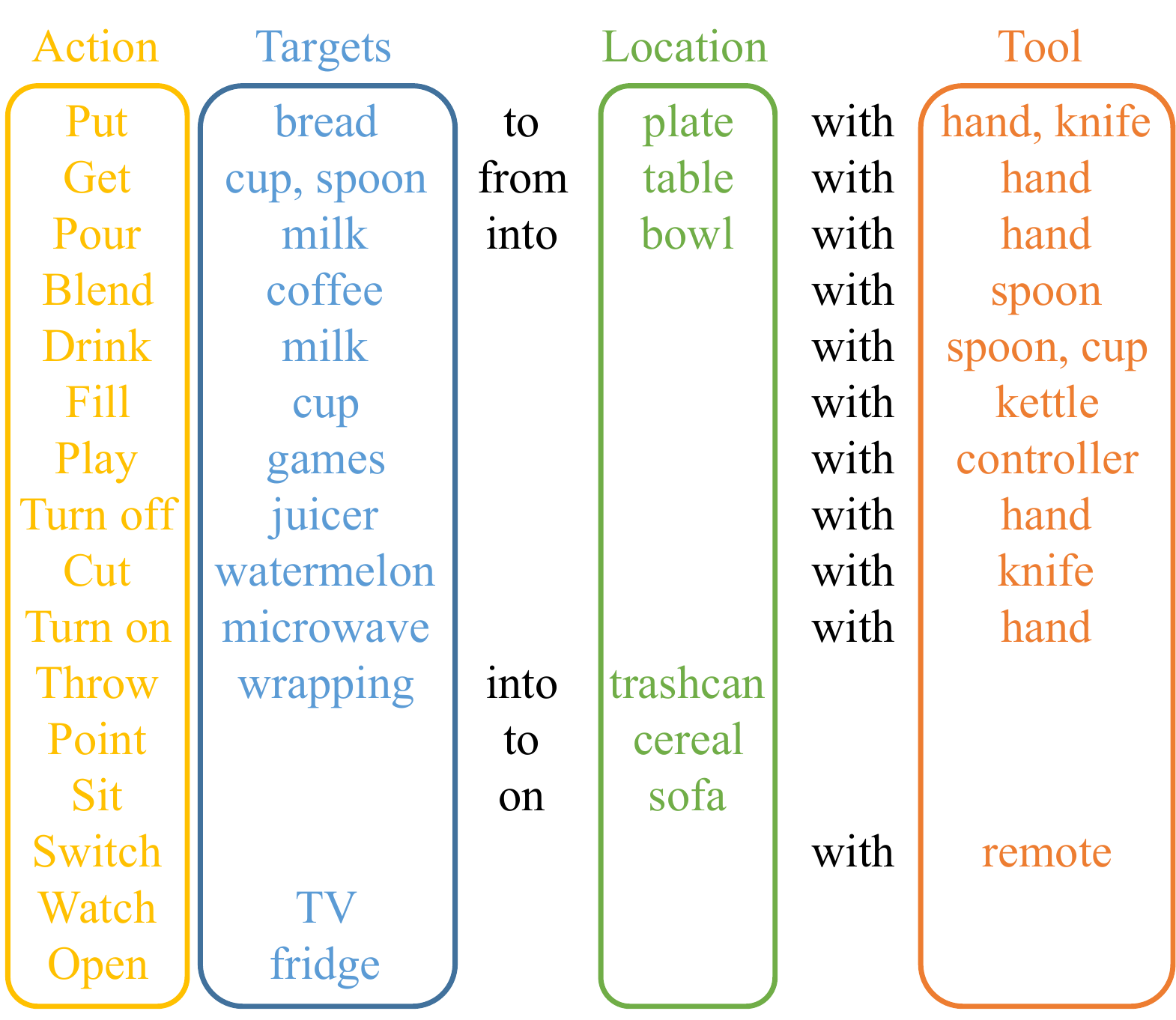}
        \caption{Compositional action templates}
        \label{fig:verb_template}
    \end{subfigure}%
    \begin{subfigure}[t]{0.54\linewidth}
        \includegraphics[width=\linewidth]{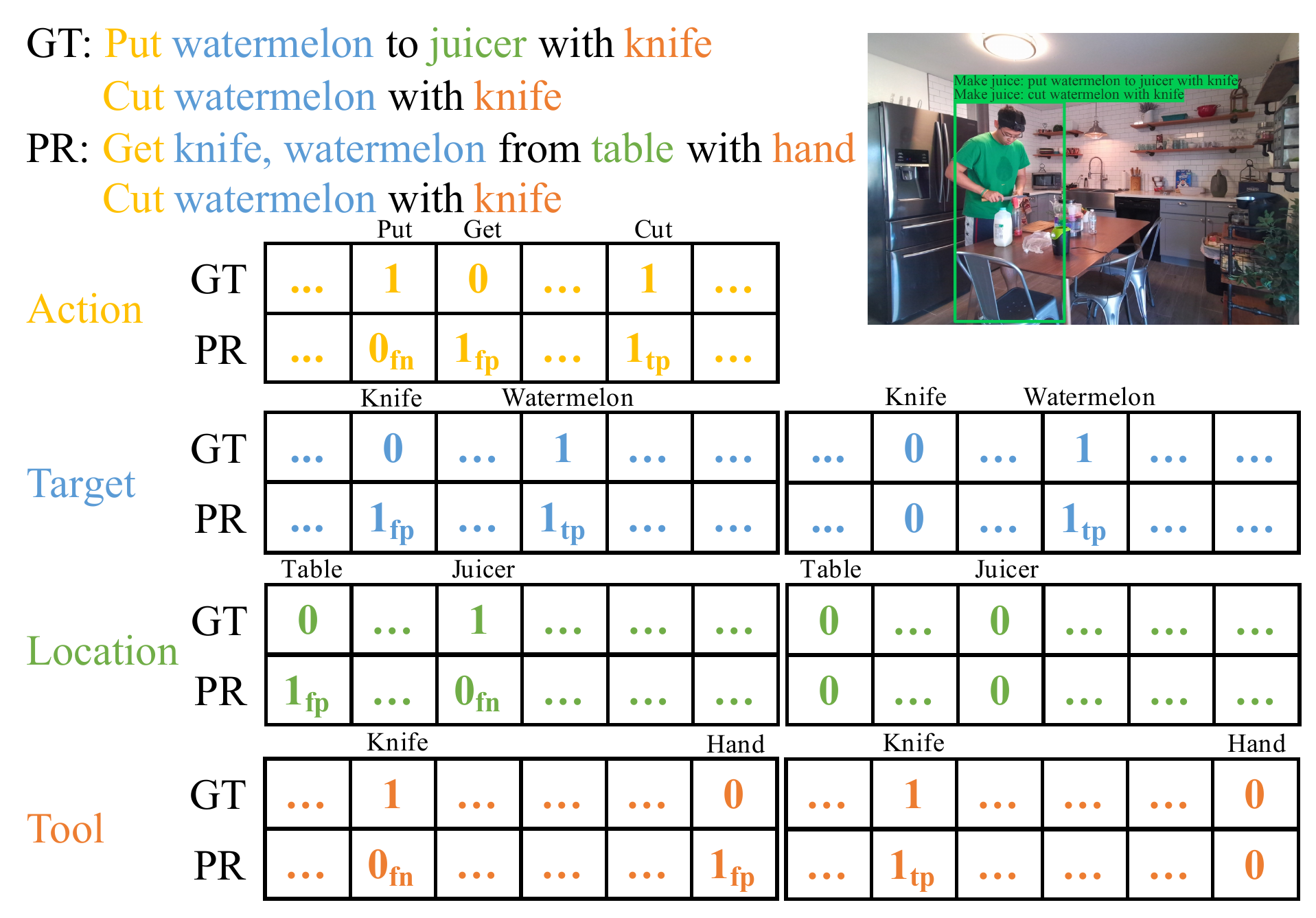}
        \caption{Prediction of verbs and nouns}
        \label{fig:composed_evaluation}
    \end{subfigure}%
    \caption{Compositional action recognition benchmark on LEMMA. (a) Examples of Compositional action templates. Yellow denotes verbs. Blue, green, and brown denote nouns for an interacting object, target/source location, and tool, respectively. (b) Examples of predictions of the verbs and nouns in compositional action recognition. Verbs and nouns are evaluated through multi-label classification.}
\end{figure*}

Motivated by the above observation, we propose the compositional action recognition benchmark on the collected LEMMA dataset with each object attributed to a specific semantic position in the action label. Specifically, we build 24 compositional action templates; see \cref{fig:verb_template} for some examples. In these action templates, each noun could denote an interacting object, a target or a source location, or a tool used by a human agent to perform certain actions.

The proposed compositional action recognition benchmark is challenging; it requires computational models to correctly detect the ongoing concurrent action verbs as well as the nouns at their correct semantic positions. We evaluate model performances by metrics on compositional action recognition in both FPVs and TPVs. Specifically, the model is asked to predict (i) multiple labels in verb recognition for concurrent actions (\eg, ``\underline{watch} tv'' and ``\underline{drink} with cup'' at the same time), and (ii) multiple labels in noun recognition for each semantic position given verbs, representing the interactions with multiple objects using the same action (\eg, ``wash \underline{spoon, cup} using sink''). \cref{fig:composed_evaluation} shows the schematics of the evaluation process. For training and testing on TPVs, we provide ground-truth bounding boxes of humans as additional information on spatial localization.

\subsection{Action and Task Anticipation}\label{sec:anticipation}

As emphasized throughout the paper, the most significant factor of human activities is the goal-directed, teleological stand. An in-depth understanding of goal-directed tasks demands a predictive ability of latent goals, action preferences, and potential outcomes. To tackle these challenges, we propose the action and task anticipation benchmark on the collected LEMMA dataset. Specifically, we evaluate model performances for the anticipation (\ie, predictions for the next action segment) of action and task with both FPV and TPV videos.

This benchmark provides both the training and testing data in all four scenarios of activities to study the goal-directed multi-task multi-agent problem. As there is an innate discrepancy of prediction difficulties among these four scenarios, we gradually increase the overall prediction difficulty, akin to a curriculum learning process, by setting the percentage of training videos to be 3/4, 1/4, 1/4, and 1/4 for $1 \times 1$, $1 \times 2$, $2 \times 1$ and $2 \times 2$ scenarios, respectively. Intuitively, with sufficient clean demonstrations of tasks in $1 \times 1$ scenario, interpreting tasks in more complex settings (\ie, $1 \times 2$, $2 \times 1$, and $2 \times 2$) should be easier, thus requiring less learning samples; such a design encourages the model to generalize. The model performance is evaluated individually for each scenario.

\section{Experiments}

In this section, we conduct experiments on the two proposed benchmarks with details on evaluation metrics, experimental settings, and baseline results. We further discuss the results to highlight the underlying challenges of each task.

\subsection{Compositional Action Recognition}\label{sec:exp:comp_rec}

\paragraph{\textbf{Experimental Setup:}}
We randomly split all the video samples into training and test sets with a ratio of 3:1, resulting in 243 recorded activities for training and the remaining 81 for testing. Due to the multi-agent setup, each activity may have multiple FPVs; 333 (out of 445) FPV videos are split into training. In TPVs, the recordings of the primary view with the ground-truth human bounding box annotations are given for both training and testing videos. Results are evaluated on two separate sources of inputs: FPVs and TPVs.

\paragraph{\textbf{Evaluation Metrics:}}
Model performances are evaluated separately for verbs, nouns, and compositional action recognition. Verb and compositional action recognition are treated as multi-label classifications with 25 verb classes and 863 compositional action classes (including a ``null'' action). After generating multi-hot labels for each semantic position in the presented verb, noun recognition is evaluated as multi-label classification (64 object classes). Average precision, recall, and F1-score for all predictions are reported on testing sets. During the evaluation, we sample image frames at 5 FPS and evaluate on these frames.

\paragraph{\textbf{Methods:}}
We adopt two recent 3D-CNN networks, I3D~\cite{carreira2017quo} and SlowFast Network~\cite{feichtenhofer2019slowfast}, as the baseline models. The baseline models predict the compositional action directly. Considering compositionality of verbs and nouns, we propose two variants of the baseline models: (i) a multi-branch network (branching model) that builds on the bottleneck layer of the backbone models to leverage both verb and noun supervision, and (ii) a multi-step inference model (sequential model), wherein verbs are first inferred with a beam search and then fed into object inference with their verb embeddings for joint learning.

\paragraph{\textbf{Implementation Details:}}
The training procedure utilizes all annotated segments in the training set. Additionally, we re-scale all the images with the short side to 256 pixels. To feed data into 3D-CNN models, 4 frames are first sampled for each action segment as center frames, and an additional 8 frames are then uniformly sampled around center frames with a window length of 32. We train each model on 8 Titan RTX GPUs on a single computing node for 50 epochs (20k iterations) with a batch size of 96. We use warm-up strategy and perform large mini-batch batch normalization, as suggested in~\cite{goyal2017accurate}. The learning rate is initially set to 0.0125 for each parallel branch and decays with a cosine annealing. Other settings of the backbone models are the same as in~\cite{feichtenhofer2019slowfast}. For the proposed sequential model, we use the beam search with a size of 5 for action inference. We extract bounding box features of humans with ROIAlign~\cite{he2017mask} for frames in TPVs. More implementation details are provided in \emph{supplementary material}.

\paragraph{\textbf{Results and Discussion:}}
\cref{tab:recognition_results} shows quantitative results of predicting verbs, nouns, and compositional actions for the compositional action recognition task. For FPVs, rather than directly predicting the compositional actions (baseline models), predicting the verbs and nouns with their semantic positions boosts the performance on all metrics, indicating that understanding the compositional structures of human actions indeed supports the prediction. We also observe that the results of compositional action recognition in the sequential models are slightly lower than the branching model due to the aggregated error brought in by a relatively low precision ($\sim$25\%) of the verb recognition. 

In comparison, the results of compositional action recognition in TPVs are significantly lower than those in the FPVs due to severe occlusion. It also shows that predicting the composition of verbs and nouns makes no significant improvement compared with predicting compositional action directly. Such a result implies that current models could not capture the details of compositions between verbs and nouns from TPVs. Taken together, the results indicate that fusion among the representations of visual embodiment between TPVs and FPVs might be a crucial ingredient to tackle this problem in the future.

\begin{figure*}[t!]
\definecolor{green(ryb)}{rgb}{0.4, 0.69, 0.2}
    \centering
    \includegraphics[width=\linewidth]{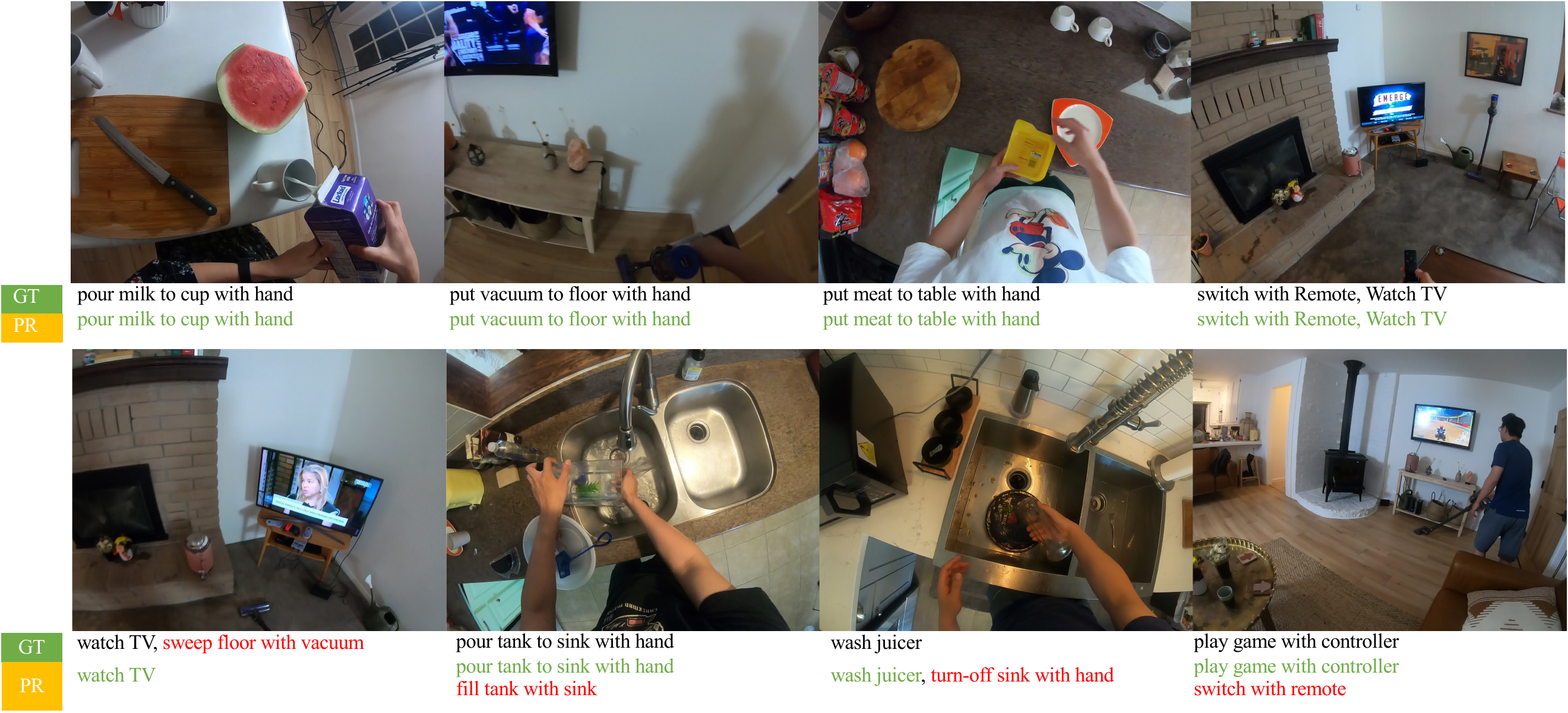}
    \caption{Qualitative results of compositional action recognition on LEMMA. From top to bottom, we show correct predictions and failure examples. {\color{red} Red} marks wrong verb or noun predictions, {\color{green(ryb)} green} indicates correct verb or noun predictions.}
    \label{fig:qualitative_result}
\end{figure*}

\cref{fig:qualitative_result} shows qualitative results for the composed action recognition task.

\begin{table}[t!]
    \centering
    \caption{Comparisons of compositional action recognition on LEMMA.}
    \label{tab:recognition_results}
    \resizebox{\hsize}{!}{%
    \begin{tabular}{c|c|c|c|c|c|c|c|c|c|c}
    \toprule
        \multirow{2}{*}{\shortstack[c]{View\\Type}} & \multirow{2}{*}{Method} & \multicolumn{3}{c|}{Verb} & \multicolumn{3}{c|}{Noun} & \multicolumn{3}{c}{Compositional Action} \\
        \cline{3-11}
        & & Avg.Prec & Avg.Rec & Avg.F1 & Avg.Prec & Avg.Rec & Avg.F1 & Avg.Prec & Avg.Rec & Avg.F1 \\
        \hline
        \multirow{6}{*}{\rotatebox[origin=c]{90}{FPV}} & I3D & 17.09 & 43.89 & 24.60 & 3.42 & 16.15 & 5.72 & 11.07 & 39.49 & 17.30 \\
        & Slowfast & 22.27 & 56.42 & 31.94 & 4.31 & 20.60 & 7.13 & 18.68 & \textbf{50.65} & 27.3 \\
        & I3D sequential & 25.04 & \textbf{57.00} & 34.80 & \textbf{19.36} & \textbf{75.29} & \textbf{30.80} & 18.00 & 50.04 & 26.47\\
        & Slowfast sequential & 24.30 & 49.71 & 32.64 & 17.95 & 59.11 & 27.54 & 26.80 & 38.41 & 31.57\\
        & I3D branching & 25.73 & 55.62 & \textbf{35.8} & 18.63 & 69.76 & 29.41 & 22.29 & 48.46 & 30.53\\
        & Slowfast branching & \textbf{26.16} & 56.33 & 35.73 & 18.18 & 73.46 & 29.15 & \textbf{27.97} & 48.87 & \textbf{35.58} \\
        \hline
        \multirow{6}{*}{\rotatebox[origin=c]{90}{TPV}} & I3D & 14.18 & 36.34 & 20.40 & 2.29 & 11.05 & 3.79 & 6.85 & \textbf{23.82} & 10.64 \\
        & Slowfast & 14.28 & \textbf{37.38} & 20.66 & 2.32 & 11.14 & 3.83 & \textbf{7.76} & 23.25 & \textbf{16.31} \\
        & I3D sequential & 16.17 & 30.17 & 21.05 & 7.79 & \textbf{25.41} & 11.93 & 2.23 & 12.67 & 3.79 \\
        & Slowfast sequential & 15.31 & 28.84 & 20.00 & 6.37 & 22.39 & 9.92 & 3.27 & 9.16 & 4.82 \\
        & I3D branching & 12.92 & 32.09 & 18.43 & 12.75 & 17.70 & 14.82 & 4.67 & 20.76 & 7.6 \\
        & Slowfast branching & \textbf{16.64} & 33.40 & \textbf{22.21} & \textbf{17.29} & 18.36 & \textbf{17.81} & 6.52 & 21.55 & 10.01 \\
        \bottomrule
    \end{tabular}
    }%
\end{table}

\subsection{Action and Task Anticipations}

\paragraph{\textbf{Experimental Setup:}}
We split the training and test sets with ratios $3 : 1$, $1 : 3$, $1 : 3$, $1 : 3$ for the four scenarios $1 \times 1$, $1 \times 2$, $2 \times 1$, $2 \times 2$, respectively. Such a spit results in training set with (96, 19, 16, 13) activities and a test set with (31, 57, 50, 42) activities in four scenarios. During training and testing, the computational models have access to both FPVs and TPVs, together with the ground-truth human bounding boxes annotations of the TPV primary view.

\paragraph{\textbf{Evaluation Metrics:}}
Model performances are evaluated individually (per agent) for the action and task anticipations task. Specifically, both action and task anticipations are evaluated as multi-label classifications with 863 compositional action classes (including a ``null'' action) and 15 task classes. Average precision, recall, and F1-score are reported individually for each of the four scenarios on the testing sets. Similar to the protocol used in the above compositional action recognition task, we re-sample image frames at 5 FPS and evaluate these sub-sampled frames during the testing phase.

\paragraph{\textbf{Methods:}}
We leverage the visual features extracted by the pre-trained SlowFast model in compositional action recognition for baseline models. Specifically, we compare two backbone models: (i) using segment-level recognition feature (SF) directly by adding an MLP on top of the features, and (ii) using long-term feature bank (LFB) with max pooling~\cite{wu2019long}.
For activities with multi-agent interactions, we use the other agent's FPV features together with their own's to capture the joint task execution progress for learning and inference; these variants are denoted as M-SF (FPV) and M-LFB (FPV).
For comparison, we also use the concatenation of the FPV feature and primary TPV feature as the input; the corresponding models are denoted as M-SF (TPV) and M-LFB (TPV).

\paragraph{\textbf{Implementation Details:}}
For the LFB model, we use a history window size of 10 and aggregate the features using max-pooling, as described in~\cite{wu2019long}. For the multi-agent variants, we use max-pooling to fuse features of two views and process them with a different branch as another temporal inference module. We train models on a single Titan Xp GPU for 50 epochs with a learning rate of 0.001. See \emph{supplementary material} for more details on network architectures.

\paragraph{\textbf{Results and Discussion:}}
\cref{tab:recognition_result2} shows quantitative results of action and task anticipation. The proposed multi-agent variants (M-) of baseline models perform the best among all models. For single-agent activities ($1 \times 1$, $1 \times 2$), we have the following crucial observations. First, models that consider temporal relations between frames generally perform better than the models using segment features. Second, adding additional TPV features to single-agent activities slightly helps interpret the task being executed and therefore promotes anticipation. This result matches the intuition that computational models having access to both FPVs and TPVs would perceive more holistic scene information. We also find that the performances of task anticipation in the $1 \times 1$ single-task scenario are better than the one in the $1 \times 2$ multi-task scenario, matching what we would expect from more complicated task execution patterns.

For multi-agent activities ($2 \times 1$, $2 \times 2$), we observe that the aggregation of FPV and TPV features generally performs better. It supports our hypothesis that observing the other agents' actions helps the computational models to ``understand'' task scheduling and assignment. We also observe that, models' performances in $2 \times 1$ activities are slightly worse than in $2 \times 2$ activities. We hypothesize that task plans in the $2 \times 2$ scenarios change less frequently, with a clear task assignment coordinates the individual tasks. In comparison, in the $2\times 1$ scenarios, the sequential ordering of the task requires more frequent communications between agents to coordinate. Such a performance gap calls for better modeling of multi-agent task assignments. Due to the page limit, we show qualitative results of action and task anticipation in the \emph{supplementary material}.

\begin{table}[t!]
    \centering
    \caption{Comparisons of the action and task anticipations on LEMMA.}
    \label{tab:recognition_result2}
    \resizebox{\hsize}{!}{%
    \begin{tabular}{c|c|c|c|c|c|c|c|c|c|c|c|c|c}
    \toprule
        \multirow{2}{*}{\shortstack[c]{Scenario}} & \multirow{2}{*}{Method} & \multicolumn{3}{c|}{$1 \times 1$} & \multicolumn{3}{c|}{$1 \times 2$} & \multicolumn{3}{c|}{$2 \times 1$} & \multicolumn{3}{c}{$2 \times 2$} \\
        \cline{3-14}
        & & Avg.Prec & Avg.Rec & Avg.F1 & Avg.Prec & Avg.Rec & Avg.F1 & Avg.Prec & Avg.Rec & Avg.F1 & Avg.Prec & Avg.Rec & Avg.F1 \\
        \hline
        \multirow{6}{*}{\rotatebox[origin=c]{90}{\shortstack[c]{Compositional\\action}}} & 
        SF & 23.42 & 22.25 & 22.82 & 20.13 & 20.06 & 20.10 & 18.89 & 19.22 & 19.05 & 18.31 & 16.67 & 17.45\\
        & LFB & 23.03 & 28.67 & 25.54 & 20.48 & 25.4 & 22.67 & 18.31 & 22.30 & 20.11 & 18.53 & 20.97 & 19.68\\
        & M-SF (TPV) & \textbf{24.22} & 28.05 & 25.99 & 20.10 & 24.48 & 22.08 & 19.15 & 16.71 & 17.85 & 19.64 & 15.18 & 17.12 \\
        & M-LFB (TPV) & 23.54 & \textbf{37.81} & \textbf{29.01} & 21.10 & \textbf{31.86} & 25.39 & 19.67 & 21.03 & 20.33 & \textbf{20.11} & 20.30 & \textbf{20.15} \\
        & M-SF (FPV) & 23.30 & 25.41 & 24.31 & \textbf{21.34} & 23.18 & 22.22 & \textbf{19.70}  & 17.46 & 18.51 & 19.82 & 15.8 & 17.58\\
        & M-LFB (FPV) & 23.26 & 31.07 & 26.60 & 20.78 & 27.40 & 23.63 & 19.42 & \textbf{21.73} & \textbf{20.51} & 19.49 & 20.12 & 19.8 \\
        \hline
        \multirow{6}{*}{\rotatebox[origin=c]{90}{Task}} 
        & SF & 50.53 & 79.08 & 61.66 & 48.07 & 67.78 & 56.25 & 39.05 & 57.43 & 46.49 & 44.88 & 62.09 & 52.1 \\
        & LFB & 57.57 & \textbf{84.31} & 68.42 & 52.12 & 68.94 & 59.36 & 38.40 & 53.08 & 44.56 & 48.17 & 64.61 & 55.19 \\
        & M-SF (TPV) & 58.61 & 79.96 & 67.05 & 55.45 & 67.24 & 60.78 & \textbf{45.73} & 58.98 & \textbf{51.51} & \textbf{49.66} & 64.47 & 56.10\\
        & M-LFB (TPV) & \textbf{60.27} & 82.19 & \textbf{69.54} & \textbf{56.2} & \textbf{72.46} & \textbf{63.30} & 43.94 & 61.41 & 51.23 & 48.85 & \textbf{67.48} & \textbf{56.67} \\
        & M-SF (FPV) & 51.12 & 79.18 & 62.13 & 48.42 & 69.04 & 56.92 & 41.00 & 58.11 & 48.08 & 46.04 & 65.97 & 54.24 \\
        & M-LFB (FPV) & 55.56 & 82.83 & 66.51 & 52.22 & 70.01 & 59.82 & 41.33 & \textbf{64.49} & 50.38 & 46.65 & 69.59 & 55.86 \\
    \bottomrule
    \end{tabular}
    }%
\end{table}

\section{Conclusions}

In this paper, we introduce the LEMMA dataset with a focus on natural multi-agent multi-task daily activities. Dense annotations are provided on both compositional action and task for learning and inference on four different activity scenarios with increasing difficulty. Additionally, we propose two challenging tasks on LEMMA to measure existing models' competence in action understanding and temporal reasoning: (i) compositional action recognition, and (ii) action/task anticipations. We hope this effort would attract the computer vision community to look into natural and realistic goal-directed human activities and further study the task scheduling and assignment in real-world scenarios.

\paragraph{\textbf{Ackowledgements:}}
We thank (i) Tao Yuan at UCLA for designing the annotation tool, (ii) Lifeng Fan, Qing Li, Tengyu Liu at UCLA and Zhouqian Jiang for helpful discussions, and (iii) colleagues from UCLA VCLA for assisting the endeavor of post-processing this massive dataset. The work reported herein was supported by ONR MURI N00014-16-1-2007, ONR N00014-19-1-2153, and DARPA XAI N66001-17-2-4029.

\clearpage
\bibliographystyle{splncs04}
\bibliography{ref}
\end{document}


\pagestyle{headings}
\mainmatter
\titlerunning{A Multi-view Dataset for \underline{L}\underline{E}arning \underline{M}ulti-agent \underline{M}ulti-task \underline{A}ctivities}

\title{Supplementary Material for\\LEMMA: A Multi-view Dataset for \underline{L}\underline{E}arning \underline{M}ulti-agent \underline{M}ulti-task \underline{A}ctivities}

\author{Baoxiong Jia \and
Yixin Chen\and 
Siyuan Huang \and
Yixin Zhu \and 
Song-Chun Zhu
}
\authorrunning{Baoxiong Jia et al.}

\institute{UCLA Center for Vision, Cognition, Learning, and Autonomy (VCLA)\\
\email{\{baoxiongjia, ethanchen, huangsiyuan, yixin.zhu\}@ucla.edu} \email{sczhu@stat.ucla.edu}}

\maketitle

\section{Annotation Details}

In this section, we describe additional details of the annotation process, including the design of the verb and noun classes, as well as the verb patterns.

\begin{table}[b!]
    \centering
    \caption{Verb vocabulary, corresponding verb patterns, and annotated examples.}
    \label{supp:tab:classes}
    \resizebox{\hsize}{!}{%
    \begin{tabular}{c|cccccc|cccccc}
         Verb &  \multicolumn{6}{c|}{Template} & \multicolumn{6}{c}{Example}\\
         \hline
         blend & blend & {\color{blue}\underline{targets}} & & & with & {\color{orange}\underline{tools}} & blend & {\color{blue}\underline{coffee}} & & & with & {\color{orange}\underline{spoon}}\\
         clean & clean & {\color{blue}\underline{targets}} & & & with & {\color{orange}\underline{tools}} & clean & {\color{blue}\underline{cup}} & & & with & {\color{orange}\underline{sink}}\\
         close & close & {\color{blue}\underline{targets}} & & & & & close & {\color{blue} \underline{drawer}}\\
         cook & cook & {\color{blue}\underline{targets}} & in & {\color{green}\underline{location}} & with & {\color{orange}\underline{tools}} & cook & {\color{blue}\underline{meat}} & in & {\color{green}\underline{pan}} & with & {\color{orange}\underline{fork}}\\
         cut & cut & {\color{blue}\underline{targets}} & on & {\color{green}\underline{location}} & with & {\color{orange}\underline{tools}} & cut & {\color{blue}\underline{watermelon}} & on & {\color{green}\underline{cutting-board}} & with & {\color{orange}\underline{knife}}\\
         drink & drink & {\color{blue}\underline{targets}} & & & with & {\color{orange}\underline{tools}} & drink & {\color{blue}\underline{milk}} & & & with & {\color{orange}\underline{cup}}\\
         eat & eat & {\color{blue}\underline{targets}} & & & with & {\color{orange}\underline{tools}} & eat & {\color{blue}\underline{meat}} & & & with & {\color{orange}\underline{fork}}\\
         fill & fill & {\color{blue}\underline{targets}} & & & with & {\color{orange}\underline{tools}} & fill & {\color{blue}\underline{juicer}} & & & with & {\color{orange}\underline{sink}}\\
         get & get & {\color{blue}\underline{targets}} & from & {\color{green}\underline{location}} & with & {\color{orange}\underline{tools}} & get & {\color{blue}\underline{spoon, fork}} & from & {\color{green}\underline{drawer}} & using & {\color{orange}\underline{hand}}\\
         open & open & {\color{blue}\underline{targets}} & & & & & open & {\color{blue}\underline{closet}}\\
         play & play & {\color{blue}\underline{targets}} & & & with & {\color{orange}\underline{tools}} & play & {\color{blue}\underline{game-console}} & & & with & {\color{orange}\underline{controller}}\\
         point-to & point to & {\color{blue}\underline{targets}} & & & & & point to & {\color{blue}\underline{kettle}} & & & & \\
         pour & pour & {\color{blue}\underline{targets}} & into & {\color{green}\underline{location}} & with & {\color{orange}\underline{tools}} & pour & {\color{blue}\underline{coffee}} & into & {\color{green}\underline{cup}} & with & {\color{orange}\underline{spoon}}\\
         put & put & {\color{blue}\underline{targets}} & to & {\color{green}\underline{location}} & with & {\color{orange}\underline{tools}} & put & {\color{blue}\underline{meat}} & to & {\color{green}\underline{pan}} & with & {\color{orange}\underline{fork}}\\
         sit-on & sit on & & & {\color{green}\underline{location}} & & & sit on & & & {\color{green}\underline{sofa}}\\
         sweep & sweep & {\color{blue}\underline{targets}} & & & with & {\color{orange}\underline{tools}} & sweep & {\color{blue}\underline{floor}} & & & with & {\color{orange}\underline{vacuum}}\\
         switch & switch & {\color{blue}\underline{targets}} & & & with & {\color{orange}\underline{tools}} & switch & {\color{blue}\underline{TV}} & & & with & {\color{orange}\underline{remote}}\\
         throw & throw & {\color{blue}\underline{targets}} & into & {\color{green}\underline{location}} & & & throw & {\color{blue}\underline{wrapping}} & into & {\color{green}\underline{trash-can}}\\
         turn-off & turn off & {\color{blue}\underline{targets}} & & & with & {\color{orange}\underline{tools}} & turn off & {\color{blue} \underline{TV}} & & & with & {\color{orange} \underline{remote}}\\
         turn-on & turn on & {\color{blue}\underline{targets}} & & & with & {\color{orange}\underline{tools}} & turn on & {\color{blue}\underline{microwave}} & & & with & {\color{orange}\underline{hand}}\\
         watch & watch & {\color{blue}\underline{targets}} & & & & & watch & {\color{blue}\underline{TV}} & & & & \\
         wash & wash & {\color{blue}\underline{targets}} & & & & & wash & {\color{blue}\underline{cup, spoon}} & & & \\
         work-on & work on & {\color{blue}\underline{targets}} & & & & & work on & {\color{blue}\underline{cup-noodles}}\\
    \end{tabular}
    }%
\end{table}

We build our action verb taxonomy and collect a compact dictionary of action verbs for annotation based on the previous dataset that captures goal-directed activities in the kitchen~\cite{rohrbach2012database,kuehne2014language,damen2018scaling} and living room~\cite{wu2015watch,savva2016pigraphs}. Specifically, we start from the action verb vocabulary summarized in the EPIC-KITCHENS dataset~\cite{damen2018scaling} to cover common actions conducted in the kitchen and living room. We then reduce the vocabulary size by eliminating unrelated action verbs, such as ``pet-down,'' ``walk,'' and ``decide-if.'' We further add action verbs (\eg, ``point to'') to incorporate human-human interactions in multi-agent collaboration scenarios. We provide this action verb vocabulary to the AMT workers for the first-round annotation. After resolving the ambiguities in language descriptions, 24 action verbs remain in the final action verb vocabulary. For the noun vocabulary, we enumerate all possible objects that could potentially be interacted in the 15 daily tasks, resulting in a noun vocabulary with a size of 64. For each action verb, we design the action verb patterns with three semantic positions for {\color{blue} interacting objects}, {\color{green} target/source location}, and {\color{orange} tools}. The action verb vocabulary and the action verb patterns are summarized in \cref{supp:tab:classes}.

\paragraph{\textbf{Bounding boxes and verbs.}}

We use Vatic~\cite{vondrick2013efficiently} to annotate the human bounding boxes and verb patterns with blanks to be filled in. The human bounding boxes are annotated as rectangles, and the verb patterns are treated as attributes for each human bounding box.

\begin{figure}[t!]
    \centering
    \includegraphics[width=\linewidth]{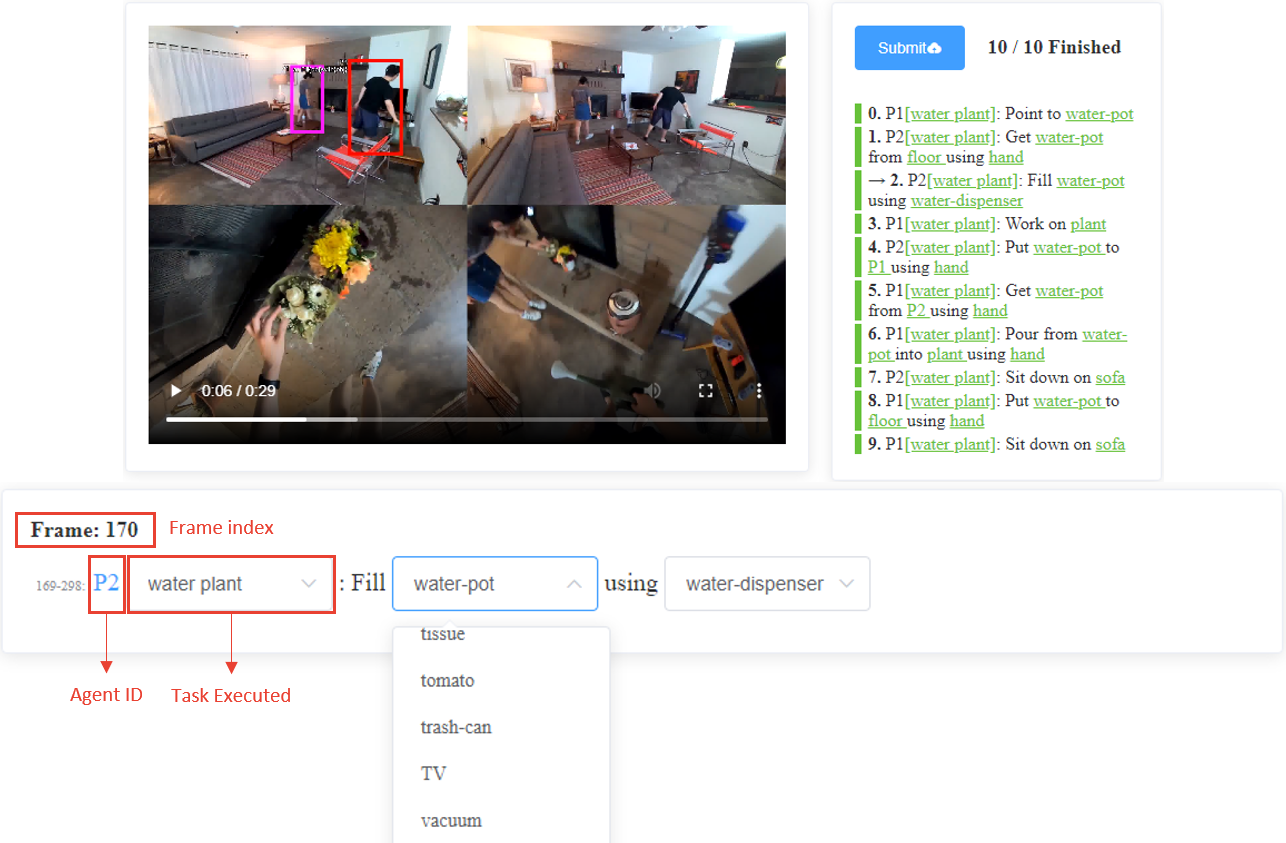}
    \caption{A visualization of the annotation tool developed for annotating the nouns and governing tasks.}
    \label{supp:fig:annotation_tool}
\end{figure}

\paragraph{\textbf{Nouns.}}

\begin{figure}[t!]
    \centering
    \includegraphics[width=\linewidth]{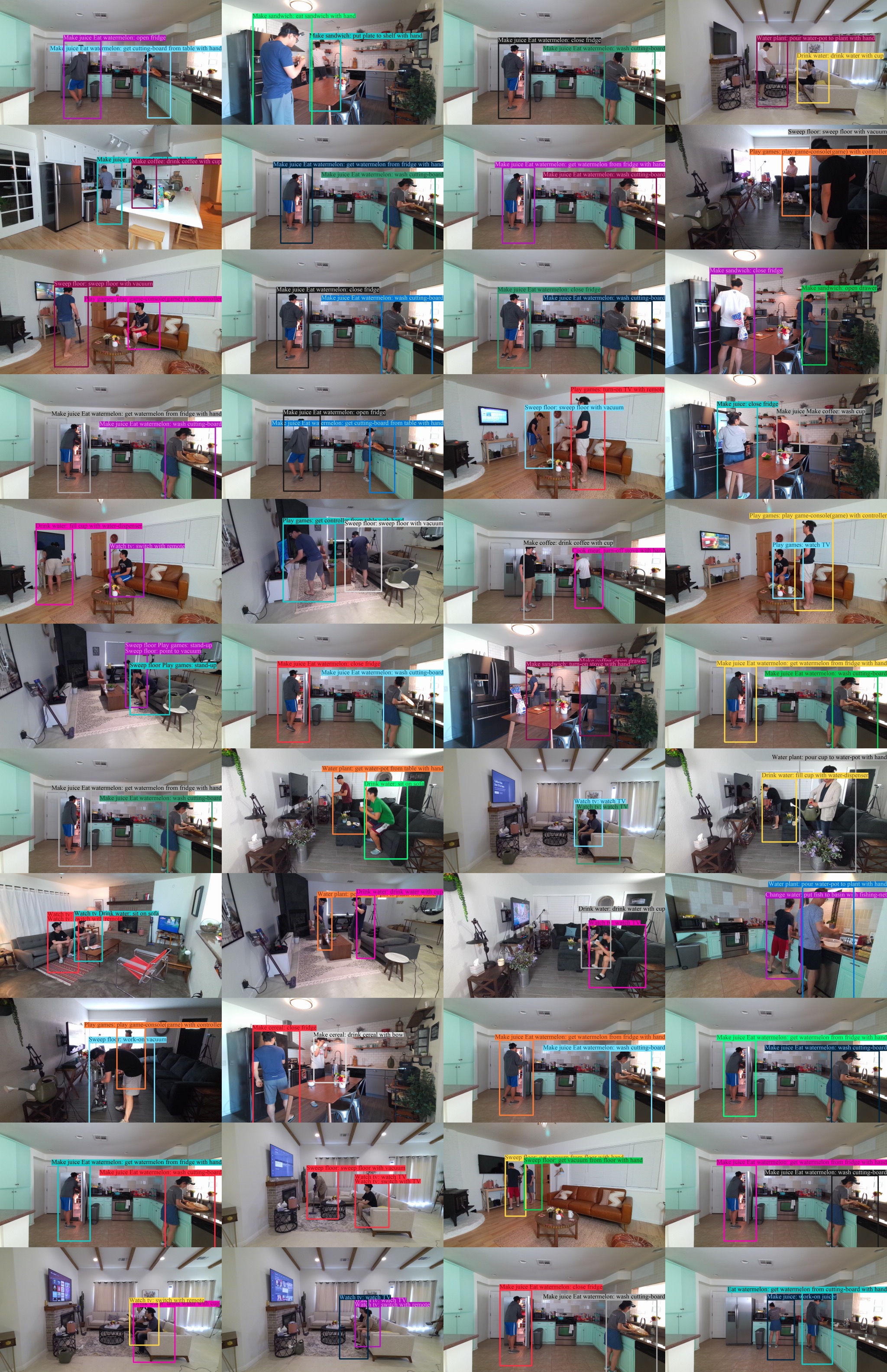}
    \caption{Examples of the annotated bounding boxes and compositional actions.}
    \label{supp:fig:data_visualization}
\end{figure}

We further develop an interactive annotation tool to fill nouns into the blanks for each semantic position of each action verb. We also use this tool to annotate the governing task of each compositional action. Each blank can be filled by choosing from a set of given options, as shown in \cref{supp:fig:annotation_tool}. During the annotation process, the synchronized video from egocentric views and TPVs are merged into the same window and presented to the AMT worker. We visualize the bounding box of agents with their ID ``P1'' and ``P2'' to help AMT workers find the correspondences. A full snapshot of the annotation interface is shown in \cref{supp:fig:annotation_tool}. After filling all blanks, we manually go through all the annotations and resolve the ambiguous action annotations by eliminating and merging the nouns with occurrence frequencies of less than 50. We show annotation results in \cref{supp:fig:data_visualization}.

\section{Implementation Details}

\paragraph{\textbf{Compositional Action Recognition}}

\begin{figure}[b!]
    \centering
    \includegraphics[width=\linewidth]{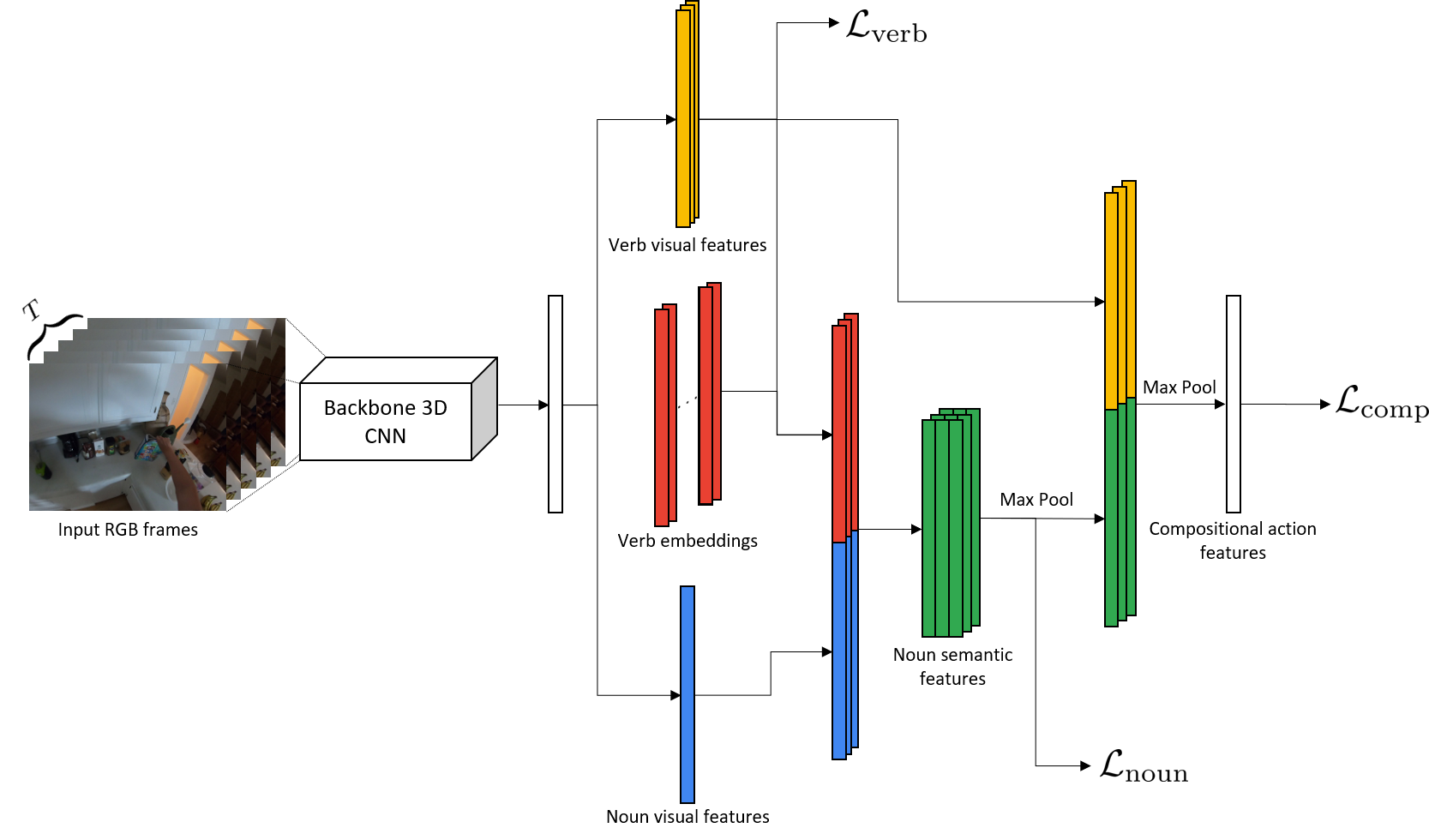}
    \caption{An illustration of the proposed ``sequential'' model, which predicts verbs, nouns, and compositional actions jointly.}
    \label{supp:fig:model_sequential}
\end{figure}

Below, we detail the designs and implementations of the two proposed models, ``branching'' and ``sequential,'' for the compositional action recognition task. We build both models on top of the backbone 3D CNN model and use a multi-branch network to train verbs, nouns, and their correspondences. We start from the ``sequential'' model as the ``branching'' model is a variant of the ``sequential'' model; see an illustration in \cref{supp:fig:model_sequential}. 

For the verb branch, we propose 3 verb candidates for each segment and extract verb visual features for verb recognition. Specifically, the verb visual features $f_{\text{verb}} = \{F_{\text{verb}}^{(i)}(f_{\text{vis}})\}$ are generated using three different linear projections $\{F_{\text{verb}}^{(i)}\}_{i=1, 2, 3}$ applied onto the feature $f_{\text{vis}}$ extracted by 3D CNN. We sort ground-truth action labels according to their index in the verb vocabulary and use cross-entropy loss $\mathcal{L}_{\text{verb}}$ as the supervision for verb recognition.

For the noun branch, we utilize the embeddings of each verb as additional features by GloVe~\cite{pennington2014glove}. The embedding of each verb is passed into a linear projection layer and concatenated with the extracted visual features to generate noun feature vectors $f_{\text{noun-vis}}$. Next, we use three different linear projections $\{F_{\text{noun}}^{(i)}\}_{i=1,2,3}$ to generate features for each of the noun visual feature vectors and obtain noun semantic features $f_{\text{noun-sem}} = \{[F^{(i)_{\text{noun}}}(f^{(j)}_{\text{noun-vis}})]_{i=1,2,3}\}_{j=1,2,3}$. As we generate ground-truth labels following the same scheme, we use binary cross-entropy loss $\mathcal{L}_{\text{noun}}$ as the supervision for recognizing nouns at their correct semantic positions using $f_{\text{noun-sem}}$. During training, the embeddings of the ground-truth verbs are fed into the network. During testing, we use the embedding of the predicted top-3 verbs.

We use max-pooling to summarize the noun semantic features and concatenate it with verb visual features. We use another layer of max pooling to generate the final compositional action feature and use binary cross-entropy loss as $\mathcal{L}_{\text{comp}}$ to provide supervision for compositional action recognition. The joint loss is
$$
\mathcal{L} = \mathcal{L}_{\text{verb}} + \mathcal{L}_{\text{noun}} + \mathcal{L}_{\text{comp}}.
$$

For the ``branching'' model, we follow the same basic scheme of the ``sequential'' model but remove the connection between the verb branch and the noun branch by discarding the additional verb embeddings. The remaining details of the architecture, as well as the optimizing objectives, remain the same.

\begin{figure}[b!]
    \centering
    \includegraphics[width=\linewidth]{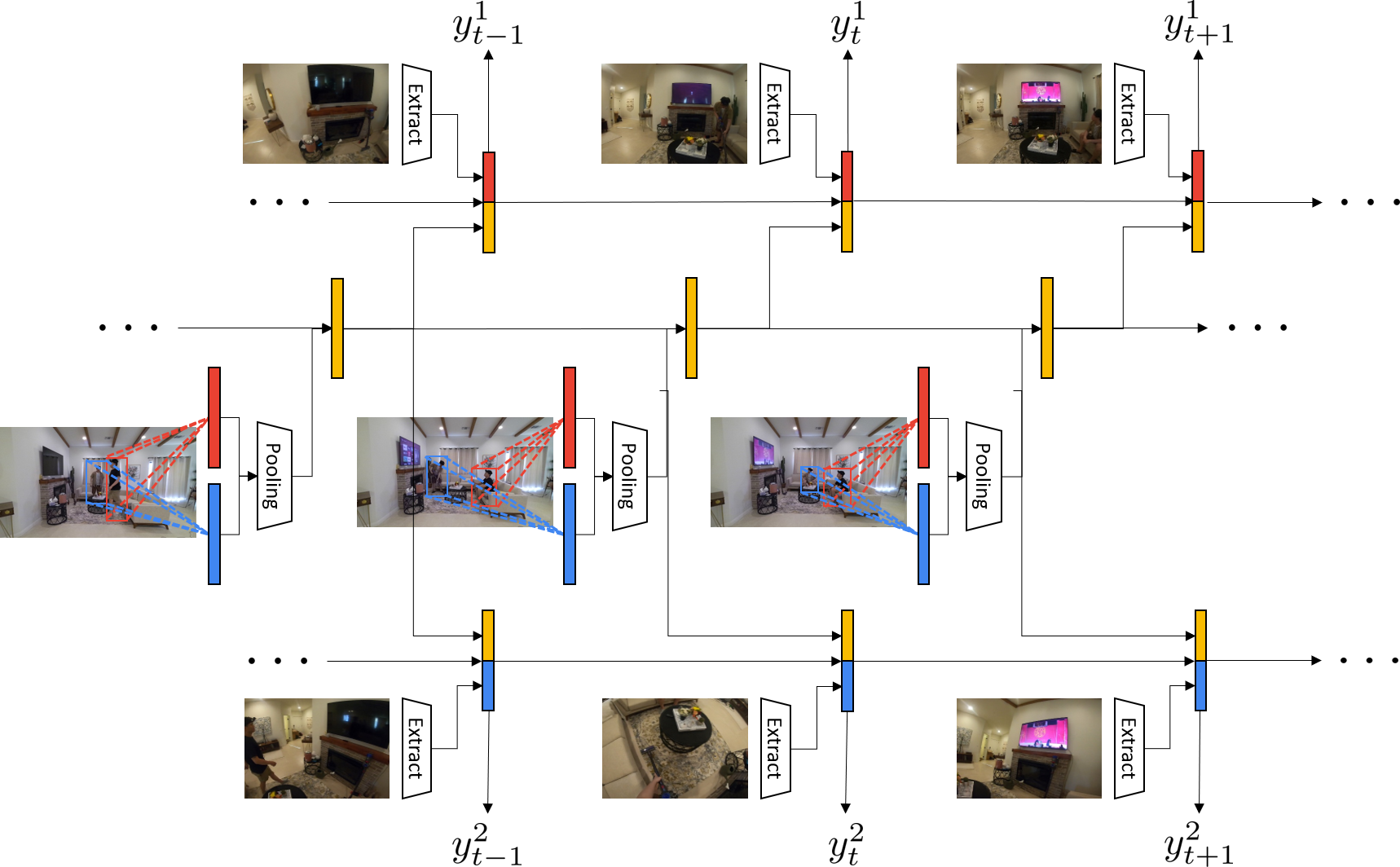}
    \caption{An illustration for the multi-agent variants of the original sequential model with TPV features as additional features.}
    \label{supp:fig:model_multi}
\end{figure}

\paragraph{\textbf{Action and Task Anticipation}}

We explain the details of the multi-agent variants of the compared sequential models. For scenarios where two agents collaborate, we incorporate the egocentric features of another agent (denoted as Ego in Table 3) or TPV features (denoted as TPV in Table 3) through a pooling mechanism, similar to~\cite{gupta2018social}. We use these pooled features to incorporate global task execution information to each agent. Specifically, we concatenate the extracted global features to features extracted by the backbone 3D CNN models from the target agent's egocentric view for training and inference. For TPV, we use ROIAlign to extract visual features corresponding to each agent's bounding box. An illustration of the pipeline with TPV features as additional features is shown in \cref{supp:fig:model_multi}.

\section{Additional Experiment Results}

We show some qualitative results for action and task anticipation.

\begin{figure}[t!]
    \centering
	\begin{subfigure}[t]{\linewidth}
		\includegraphics[width=\linewidth]{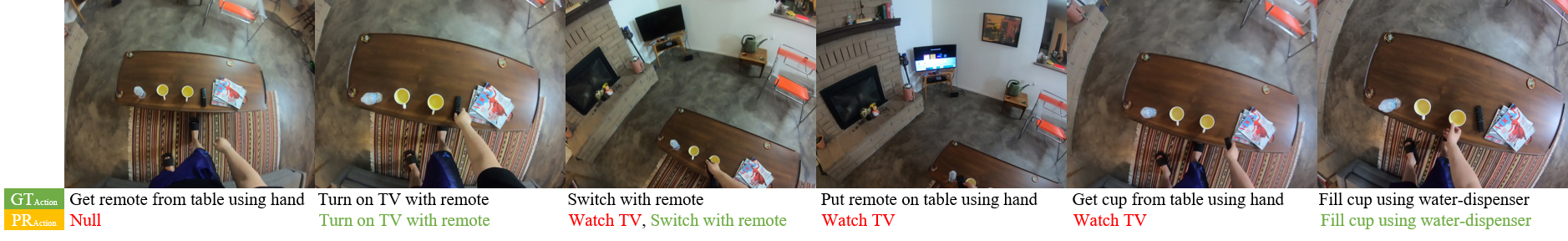}
		\caption{Results of action anticipation. GT\textsubscript{Action} indicates the ground-truth action in the next frame.}
		\label{supp:fig:anicipation_action}
	\end{subfigure}%
	\\
	\begin{subfigure}[t]{\linewidth}
		\includegraphics[width=\linewidth]{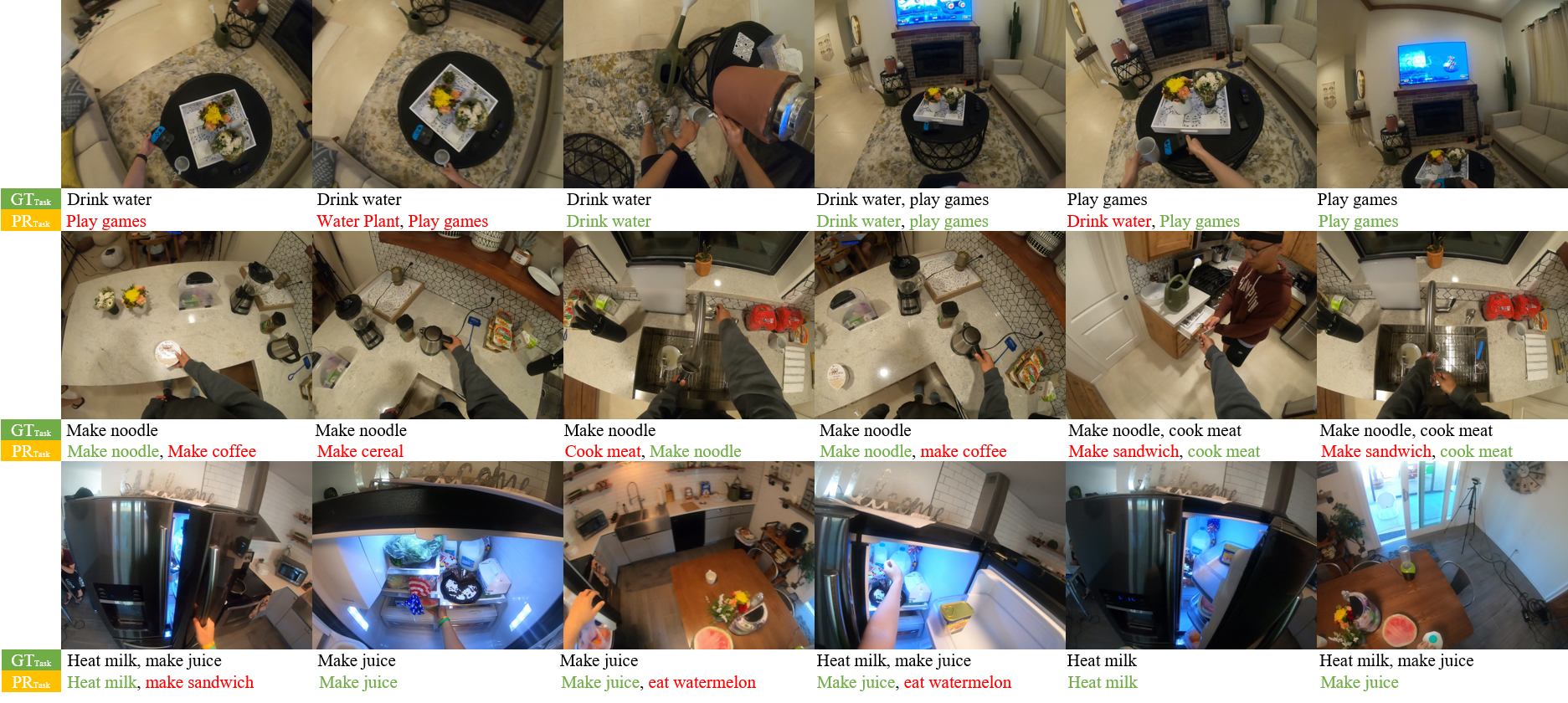}
		\caption{Results of task anticipation. GT\textsubscript{Task} indicates the ground-truth task for actions in the next segment.}
		\label{supp:fig:anicipation_task}
	\end{subfigure}%
	\caption{Qualitative results of action and task anticipation on LEMMA.}.
    \label{supp:fig:anticipation}
\end{figure}

\clearpage
\bibliographystyle{splncs04}
\bibliography{ref_bib}